%% file: main.tex
\documentclass[runningheads]{llncs}

\usepackage{eccv}

\usepackage{eccvabbrv}

\usepackage{graphicx}
\usepackage{booktabs}
\usepackage{multirow}
\usepackage{algorithm}
\usepackage{algorithmic}
\usepackage{pifont}
\usepackage{amsmath,amssymb}
\usepackage{xcolor}
\definecolor{lightred}{RGB}{220,80,80}
\usepackage{subcaption}
\usepackage{wrapfig}

\newcommand{\cmark}{\ding{51}}
\newcommand{\xmark}{\ding{55}}

\newcommand{\ours}{RayRoPE}

\usepackage[accsupp]{axessibility}  % Improves PDF readability for those with disabilities.

\usepackage{hyperref}

\usepackage{orcidlink}

\begin{document}

% ---------------------------------------------------------------
\title{\ours: Projective Ray Positional Encoding for Multi-view Attention}
\titlerunning{\ours: Projective Ray Positional Encoding}

\author{
Yu Wu\inst{1} \and
Minsik Jeon\inst{1} \and
Jen-Hao Rick Chang\inst{2} \and
Oncel Tuzel\inst{2} \and
Shubham Tulsiani\inst{1}
}

\authorrunning{Y.~Wu et al.}
% If there are more than two authors, 'et al.' is used.

% TODO FINAL: Replace with your institution list.
% \institute{
% Carnegie Mellon University \and Apple
% }
\institute{
Carnegie Mellon University \and Apple
}
\maketitle

{\centering
{\small \textbf{Project Page:} \href{https://rayrope.github.io/}{\textcolor{lightred}{https://rayrope.github.io/}}}\\
\par}

\input{sec/0_abstract}

\input{sec/1_intro}

\input{sec/2_relevant_work}

\input{sec/3_preliminaries}

\input{sec/4_approach}

\input{sec/5_experiments_reorg}

\input{sec/6_discussion}

\clearpage
\bibliographystyle{splncs04}
\bibliography{main}

\clearpage
\input{sec/X_suppl}

% ---- Bibliography ----
%
% BibTeX users should specify bibliography style 'splncs04'.
% References will then be sorted and formatted in the correct style.
%
\end{document}

%% file: sec/0_abstract.tex
\begin{abstract}
% ---------- Previous Version ---------- 
% We study positional encodings for multi-view transformers that process tokens from a set of posed input images, and seek a mechanism that encodes patches uniquely, allows $SE(3)$-invariant attention with multi-frequency similarity, and can be adaptive to the geometry of the underlying scene. We find that prior (absolute or relative) encoding schemes for multi-view attention do not meet the above desiderata, and present \ours~to address this gap. \ours~represents patch positions based on associated rays but leverages a predicted point along the ray instead of the direction for a geometry-aware encoding. To achieve $SE(3)$ invariance, \ours~computes query-frame projective coordinates for computing multi-frequency similarity. Lastly, as the `predicted' 3D point along a ray may not be precise,  we present a mechanism to analytically compute the expected position encoding under uncertainty. We validate \ours~on the tasks of novel-view synthesis and stereo depth estimation and show that it consistently improves over alternate position encoding schemes (\eg $15\%$ relative improvement on LPIPS in CO3D). 
% removed the below sentence
% We also show that \ours~can seamlessly incorporate RGB-D input, resulting in even larger gains over alternatives that cannot positionally encode this information. 

% ----------  New version March 5th ---------- 
We study positional encodings for multi-view transformers that process tokens from a set of posed input images, and seek a mechanism that encodes patches uniquely, allows $SE(3)$-invariant attention with multi-frequency similarity, and can adapt to the geometry of the underlying 3D scene. We find that prior (absolute or relative) encoding schemes for multi-view attention do not meet these desiderata, and present \ours~to address this gap. \ours~represents patch positions based on associated rays and computes query-frame projective coordinates to ensure $SE(3)$ invariance. To adapt to scene geometry, \ours{} predicts (without direct supervision) a per-token depth to obtain its position along the corresponding ray, while also modeling uncertainty and analytically computing the expected positional encoding. We validate our method on the tasks of novel-view synthesis, stereo depth estimation, and feed-forward 3DGS reconstruction. While remaining efficient, \ours{} consistently improves over alternate position encoding schemes (\eg $24\%$ relative improvement on LPIPS in RE10K).

% ---- For text abstract --------
% We study positional encodings for multi-view transformers that process tokens from a set of posed input images, and seek a mechanism that encodes patches uniquely, allows SE(3)-invariant attention with multi-frequency similarity, and can adapt to the geometry of the underlying 3D scene. We find that prior (absolute or relative) encoding schemes for multi-view attention do not meet these desiderata, and present RayRoPE to address this gap. RayRoPE represents patch positions based on associated rays and computes query-frame projective coordinates to ensure SE(3) invariance. To adapt to scene geometry, RayRoPE predicts (without direct supervision) a per-token depth to obtain its position along the corresponding ray, while also modeling uncertainty and analytically computing the expected positional encoding. We validate our method on the tasks of novel-view synthesis and stereo depth estimation. While remaining efficient, RayRoPE consistently improves over alternate position encoding schemes (e.g., 24% relative improvement on LPIPS in RE10K and 15% in CO3D).

\keywords{positional encoding \and multi-view attention \and transformer}
\end{abstract}

%% file: sec/1_intro.tex
\section{Introduction}
\label{sec:intro}

\begin{figure*}[t]
  \centering
  % \fbox{\rule{0pt}{10cm} \rule{0.95\linewidth}{0pt}}
  \includegraphics[width=1.0\textwidth]{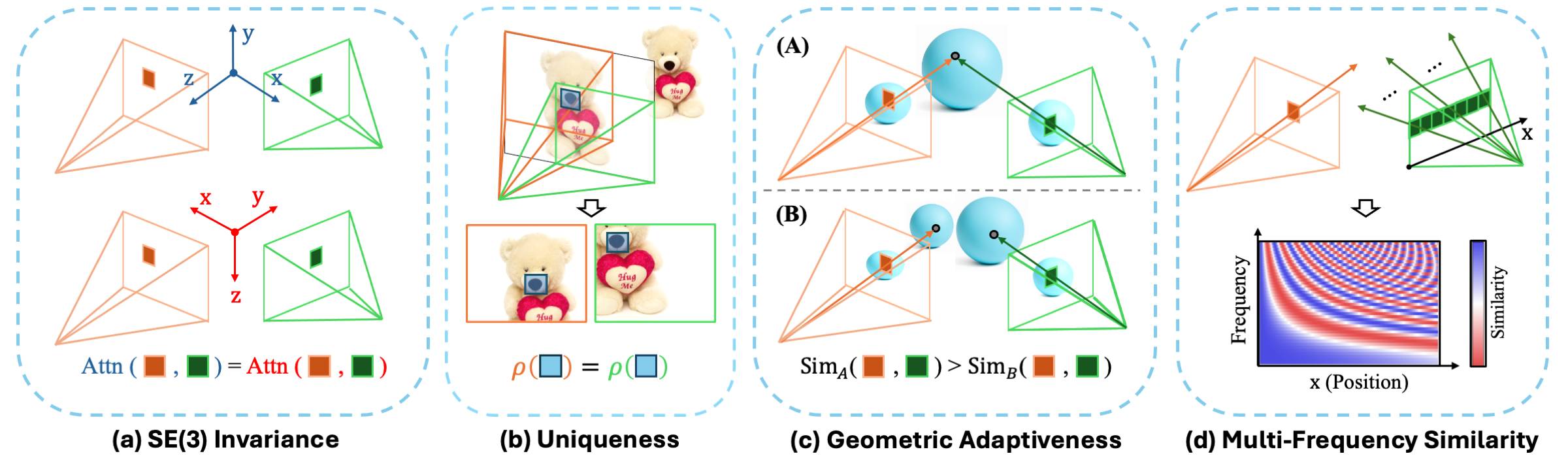}
  \caption{\textbf{Desiderata for a Multi-View Positional Encoding.} We seek the following properties for positional encoding in multi-view attention: \textbf{(a)} The attention output should be invariant to the choice of global coordinate system (i.e., $SE(3)$ invariance). \textbf{(b)} The positional encoding of tokens that correspond to the same patch observed across different images should be the same. \textbf{(c)} The positional encoding can vary with the underlying scene geometry (\eg yielding higher similarity when patches observe a common 3D point than when they do not). \textbf{(d)} Analogous to common 1D and 2D encodings, aspects of the positional encoding should vary at different frequencies, enabling multi-frequency similarity. \ours satisfies (a--d) by encoding ray segments projected into each query camera frame.}
  \label{fig:teasor}
\end{figure*}
% oh I see. thanks!

% Para 1 (framing the question we focus on): Transformers provide a general mechanism for `tokens' to attend to one another, with applications in language, image processing etc. In this work we focus on settings where the tokens correspond to patches from multi-view images \eg novel view synthesis, stereo synthesis, or camera controlled generation. The specific question we consider is how we should represent the positions of these tokens?
Vision transformers have become ubiquitous~\cite{vit,clip,dino,dinov3} in image processing applications. At their core is a simple but general computation mechanism where image patches are encoded as `tokens' and update their representations by `attending' to one another~\cite{transformer}. This attention operation does not natively account for the position of the input tokens (\eg whether a patch lies in the top left or the bottom right), so it is typical to additionally use a positional encoding to impart this information. When the tokens correspond to patches from a single image, pixel space serves as a natural coordinate system, and we can define positional encodings as a function of the 2D patch position. However, such an encoding is not always applicable: for tasks such as novel-view synthesis or multi-view stereo, tokens can originate from \emph{different} images and cannot be associated with a single 2D coordinate system. In this work, we consider \emph{multi-view} vision transformers and ask: \emph{how should we define positional encodings for tokens corresponding to patches from a set of posed images?}

% Para 2 (on position embeddings and the typical approach for MV settings is just analogous to APE): A transformer layer in position equivariant, and thus position encodings are critical to inform about the structure of inputs. The classical approach is to use APEs, but recently, the community has switched to relative ones \eg RoPE. For multi-view settings however, the popular choice is to just encode plucker rays in some but this is analogous to APEs. In this work, we investiagte whether we can design RoPE analogue.

We seek to formulate a mechanism with the following properties:
%In particular, we seek to formulate a position encoding mechanism for multi-view transformers that exhibits the following properties --
% However, we argue that a position encoding mechanism for multi-view vision transformers should exhibit the following --
\textbf{\emph{(a) $SE(3)$ invariance}}: the attention operation should only depend on \emph{relative} cameras and not a global coordinate system, \textbf{\emph{(b) uniqueness}}: if the same patch is viewed across different images (\eg overlapping crops of a panorama), the position of the corresponding tokens should be the same, \textbf{\emph{(c) geometry-adaptiveness}}: position encodings can vary with underlying scene geometry \eg if different patches `see' the same corresponding 3D point, their positions should be more similar than if the same patches see different 3D points, and \textbf{\emph{(d) multi-frequency similarity}}: as typical for 1D and 2D position encodings~\cite{fourier}, a multi-view position encoding should also compute similarity in attention at various `frequencies'. While we are certainly not the first to consider the question of designing position encodings for multi-view transformers~\cite{cat3d,zero123,ac3d}, current formulations~\cite{cape,gta,prope} do not meet the above desiderata. In particular, these methods define position encodings via image coordinates and camera matrices, and while they ensure $SE(3)$ invariance, they do not represent patches uniquely, encode positions independently of (known or estimated) scene geometry, and do not leverage the camera information in a `multi-frequency' manner.

We present a formulation, \ours, that leverages the ray(s) corresponding to each patch to define positional embeddings for multi-view transformers. While this directly yields unique, multi-frequency encodings, these are not $SE(3)$ invariant and make attention depend on an arbitrary global coordinate system. \ours~overcomes this by instead using \emph{relative} ray positions for attention, transforming them into the (projective) coordinates of the query token's camera. Additionally, instead of parameterizing rays with their origin and direction (equivalently, a point at infinity), \ours~replaces the latter by letting each token \emph{predict} (without direct supervision) a point along the ray to encode its position, allowing the embedding to vary with the underlying geometry. We also formulate a mechanism to efficiently incorporate uncertainty in predicted depths along a ray, yielding an analytical solution to compute the expected positional encoding across frequencies. Our formulation can be implemented efficiently, with runtime comparable to existing positional encoding methods.

We validate \ours~on a variety of 3D vision tasks, including novel-view synthesis~\cite{lvsm,cape}, stereo depth estimation~\cite{unimatch}, and feed-forward 3D Gaussian Splatting\cite{3dgs} reconstruction~\cite{tokengs}. We integrate our positional encoding scheme into prior methods that leverage multi-view transformers for various tasks, and show that \ours~outperforms all prior (relative and absolute) positional encodings.
%-- implementing a simple multi-view transformer that takes as input some `tokens' from posed reference images and (unobserved) target images, and updates these via multi-view attention to predict target views, and b) stereo depth estimation -- integrating our position encoding in a multi.
We then perform extensive ablations and analysis on \ours, verifying the importance of the above desiderata in our method and showing strong generalization beyond training views and datasets. We also show that \ours's `depth prediction' in the attention mechanism allows geometry to emerge without direct supervision.
% and also show that \ours~can effectively integrate known geometry at inference (\eg RGB-D reference views in novel view synthesis).

Our core contributions can be summarized as follows:
\begin{itemize}
    \item We propose \ours \, a first multi-view positional encoding satisfying the desiderata above: uniqueness, invariance, geometric adaptiveness, and multi-frequency encodings.
    \item Our formulation enables tokens to determine their own positions via (emergent) depth and extends RoPE to model uncertainty in these positions.
    \item We provide an efficient and general implementation of \ours{} which can be easily integrated into generic multi-view models, improving performance over prior positional encoding schemes.
\end{itemize}

%% file: sec/2_relevant_work.tex
\section{Related Work}
\label{sec:relevant-work}

% try emphasize the progress from APE to RPE
% we used to APE, now people move on to RPE
% also introduce frequency here

\textbf{Positional Encodings in Transformers.}
The attention mechanism is inherently permutation-invariant, treating its input tokens as an unordered set~\cite{transformer}. To recover structural information, transformers require explicit positional encodings. Early approaches~\cite{bert,sam,dino,latentdiffusion,clip,vit} relied on absolute positional encodings~(APE), which add or concatenate fixed or learned positional embeddings to token representations. More recently, the community shifted toward relative positional encodings~(RPE), which model relative positions between token pairs by modifying the attention mechanism. This includes additive attention biases~\cite{tttransformer,trainshort,relpose} and rotary positional encodings~(RoPE)~\cite{rope}, which rotate query and key vectors in a position-dependent manner. RoPE confers translational invariance to transformers and has progressively become the standard positional encoding approach adopted in leading models across diverse domains~\cite{tttransformer,llama,deepseek,visualinstruction,hunyuanvideo,string,2drope}. We propose a RoPE-based positional encoding tailored for multi-view transformers, designed to encode spatial relationships in a manner invariant to arbitrary rigid transformations of the coordinate system.

\noindent\textbf{Multi-view Transformers.}
Multi-view transformers have emerged as a powerful framework driving progress across a wide range of 3D vision tasks, such as novel-view synthesis~\cite{cat3d,lvsm,lrm,gs-lrm,seva,ac3d,bolt3d} and 3D scene understanding~\cite{input-3d-bias,zerodepth,3dconcept,odin}. These models take as input a set of posed images, where the key challenge is to encode spatial relationships among image patches across views. Existing approaches typically represent camera information as rays~\cite{cat3d,lvsm,seva,zerodepth} or camera matrices~\cite{zero123,input-3d-bias} and concatenate them with input features, analogous to absolute positional encodings (APE). However, these encodings rely on an arbitrary global coordinate system; our method instead ensures $SE(3)$ invariance via relative encoding.

% these methods depend on a fixed world coordinate system, imposing an arbitrary global reference frame that can limit generalization and robustness across diverse scenes. 
% In contrast, our work aims to develop a positional encoding analogous to relative positional encodings (RPE), enabling invariance to rigid transformations of the global coordinate frame.

\noindent\textbf{Relative Positional Encodings in 3D.}
Closest to our work are relative positional encodings based on camera poses~\cite{cape,gta,prope,viewrope}. Designed for multi-view transformers, these methods transform attention features using camera pose matrices and ensure invariance to the choice of global coordinate system (see Sec.~\ref{sec:camera-rpe}). Recent work \cite{kaleido,bullettime,viewrope,rerope} also demonstrates that such pose-based relative encodings can be applied to video generation models for camera control. Despite their effectiveness, they cannot adapt to scene geometry or support multi-frequency similarity as in standard RoPE. While some methods~\cite{gta, prope} alleviate this by incorporating standard RoPE on patch indices, such a design breaks uniqueness and is not geometrically grounded across views. In comparison, we propose a ray-based relative positional encoding that naturally supports frequency modulation, ensures uniqueness, and can adapt to the geometry of the underlying 3D scene being observed. Concurrent work~\cite{penc-field} augments standard RoPE with depth, but its generalizability to settings with more than two views is not explored.

% move thing around between Section 2 and 3.1 preliminaries?

% in section 3
% introduce notations
% first list 3 desirable properties
% analyze PRoPE, it doesn't satisfy

%% file: sec/3_preliminaries.tex
\section{Preliminaries: Positional Encodings}
\label{sec:preliminaries}

We consider the attention mechanism operating on a set of token features $\{ \boldsymbol\tau_i \}_{i=1}^L$. We use $\mathbf{q}_i, \mathbf{k}_i, \mathbf{v}_i\in \mathbb{R}^D$ to denote the corresponding query, key, and value feature of token $\boldsymbol\tau_i$. Consider a query token $\boldsymbol\tau_i$ that attends to a set of tokens $\{\boldsymbol\tau_j\}$, the output of attention on token $\boldsymbol\tau_i$ can be written as:
\begin{equation}
    \mathbf{o}_i = \text{Attn}(\mathbf{q}_i, \{\mathbf{k}_j,\mathbf{v}_j\})= \frac{\sum_j \exp(\mathbf{q}_i^\intercal \mathbf{k}_j)\mathbf{v}_j}{\sum_j \exp(\mathbf{q}_i^\intercal \mathbf{k}_j)}
    \label{eq:attention}
\end{equation}
The pairwise nature of attention allows the design of relative positional encodings that provide certain invariance properties. In the following sections, we review the standard Rotary Position Embeddings (RoPE) (Sec.~\ref{sec:rope}) for translational invariance and recent camera-based relative positional encodings (Sec.~\ref{sec:camera-rpe}) for $SE(3)$ invariance. We then analyze the limitations of these approaches (Sec.~\ref{sec:4properties}).

\subsection{Rotary Positional Encoding}
\label{sec:rope}
% shorter
RoPE was initially proposed to encode the 1D positions $x_i$ in language models by transforming features with a series of SO(2) rotations at various frequencies. We use $\oplus$ to denote matrix concatenation along the diagonal. RoPE encodes a position $x$ as a $D \times D$ matrix:

\begin{equation}
    \rho_D(x) \equiv \oplus_{f=1}^{D/2} \rho_2(\omega_f x) \equiv \oplus_{f=1}^{D/2} e^{i\omega_fx}
    \label{eq:rope}
\end{equation}
where we denote $2 \times 2$ rotational matrices with $e^{i\omega x}$ for notational simplicity. $\{\omega_f\}_{f=1}^{D/2}$ is a set of predefined rotational frequencies. RoPE encoding is used to transform query and key features, yielding $\rho(x_i)\mathbf{q}_i, \rho(x_j)\mathbf{k}_j$. The resulting attention score between any two tokens only depends on the relative position $x_j-x_i$, making the attention invariant to translations of positions. By rotating different channels of features at various frequencies, the model can learn to reason about the positional information at different scales (see Fig.~\ref{fig:teasor}~(d)).
For a general position $\mathbf{x}\in \mathbb{R}^{C}$ that is $C$-dimensional, it is straightforward to extend RoPE:
\begin{equation}
    \rho_D(\mathbf{x}) \equiv \oplus_{f=1}^{D/2C} \oplus_{c=1}^{C} e^{i\omega_fx_c}
    \label{eq:nd-rope}
\end{equation}
This is commonly used for the pixel indices $(u,v)$ in vision transformers~\cite{2drope,sam2,dino,dinov3} and the three-dimensional positions $(u,v,t)$ in video transformers~\cite{cogvideox,hunyuanvideo}. For multi-view transformers, while it is possible to directly apply RoPE to positions in 3D (such as ray maps), the relative positions will depend on the rotation of the global coordinate frame, leading to suboptimal performance (see Sec.~\ref{sec:exp-nvs}).
% However, this method is only provides invariance to translation but not rotation, which is neccessary for applications in 3D.

\subsection{Camera-Based Relative Positional Encoding}
\label{sec:camera-rpe}

Instead of using SO(2) encodings, \textbf{CaPE}~\cite{cape} encodes cameras using their extrinsics $T\in \mathbb{R}^{4\times4}$. The corresponding $D \times D$ encoding is computed by repeating $T$ along the diagonal, denoted as $E^{\text{cape}}_i =\oplus_{n=1}^{D/4} T_i$. The resulting attention score only conditions on the relative poses $T_iT_j^{-1}$, which is independent of the choice of global coordinate frame, making the transformer model $SE(3)$ invariant.
However, the camera-based encoding used in CaPE cannot support multi-frequency similarities, limiting the model's ability to reason about high-frequency details. \textbf{GTA}~\cite{gta} compensates for this by incorporating standard RoPE on image patch indices. A patch position is represented by $\mathbf{x}=(T,u,v)$, and the corresponding encoding is a concatenation of CaPE and standard 2D RoPE:
\begin{equation}
    E^{\text{gta}}_i = E^{\text{gta}}(\mathbf{x}_i) \equiv (\oplus_{n=1}^{D/8} T_i) \oplus \rho_{\frac{D}{2}}(u_i,v_i)
    \label{eq:gta}
\end{equation}
In addition to applying encodings to query and key features, GTA also applies encodings to value and output features. 
% It alters the attention mechanism in Eq.\ref{eq:attention} into:
% \begin{equation}
% \begin{aligned}
% &\text{Attn}^{\text{gta}}(\mathbf{q}_i, \{\mathbf{k}_j,\mathbf{v}_j\}) =\\
% &\quad E_i \text{Attn}(E_i^\intercal \mathbf{q}_i, 
%     \{E_j^{-1}\mathbf{k}_j, 
%     E_j^{-1}\mathbf{v}_j\})
% \end{aligned}
% \label{eq:gta-attn}
% \end{equation}
\textbf{PRoPE}~\cite{prope} further improves GTA by replacing the camera extrinsics $T$ with the full projection matrix $P=KT \in \mathbb{R}^{4\times4}$ (where the intrinsics $K$ are lifted to $4\times 4$). This enables the model to also reason about the camera intrinsics.

\subsection{Limitations of Existing Methods}
\label{sec:4properties}
In Fig.~\ref{fig:teasor}, we illustrate four desirable properties for positional encodings in multi-view transformers: \textbf{(1)} $SE(3)$ invariance, \textbf{(2)} uniqueness, \textbf{(3)} geometry-adaptiveness, and \textbf{(4)} multi-frequency similarity. While standard RoPE supports multi-frequency encoding, it is not $SE(3)$ invariant. The camera-based relative encodings~\cite{cape,gta,prope} satisfy $SE(3)$ invariance, but cannot perform multi-frequency encodings. GTA and PRoPE compensate for this by including standard 2D RoPE based on patch indices $(u,v)$. This hybrid design, however, still does not incorporate frequencies for the camera and also breaks the uniqueness property. As illustrated in Fig.~\ref{fig:teasor}~(d), if the same image patch is viewed across different overlapping images, its patch indices $(u,v)$ will change, resulting in a different positional encoding. Furthermore, these prior methods lack explicit geometry adaptiveness, and the computed encodings cannot vary with the underlying 3D scene structure (Fig.~\ref{fig:teasor}~(c)). % Consequently, they cannot ensure that patches observing the same 3D point have more similar encodings.

%% file: sec/4_approach.tex
\begin{figure*}[t]
  \centering
  % \fbox{\rule{0pt}{10cm} \rule{0.95\linewidth}{0pt}}
  \includegraphics[width=1.0\textwidth]{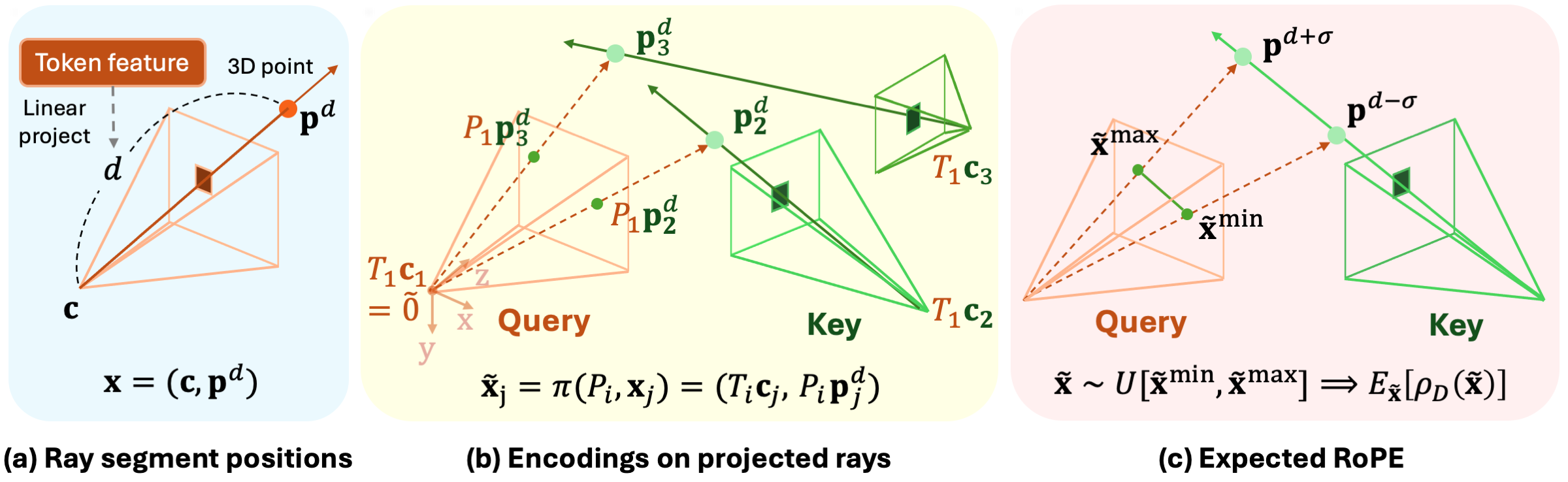}
  \caption{\textbf{Overview of \ours.} \textbf{(a)} We encode image patch position as a ray segment $\mathbf{x}=(\mathbf{c},\mathbf{p}^d)$, where $\mathbf{c}$ is the camera center and $\mathbf{p}^d$ is the point at depth $d$ along the ray $r$. We use a linear layer to allow each token to predict the depth $d$ along the ray, thus enabling  \ours{} to adapt to the scene geometry.
  \textbf{(b)} To ensure $SE(3)$ invariance, we compute the positional encodings using ray positions projected to the query camera frame with $P_i=K_iT_i$, yielding $\tilde{\mathbf{x}}_j=\pi(P_i,\mathbf{x}_j)$.  
  \textbf{(c)} To model the uncertainty in depth prediction, we also predict an uncertainty $\sigma$, yielding an estimated range between $\mathbf{p}^{d-\sigma}$ and $\mathbf{p}^{d+\sigma}$, and use an analytically computed expected position encoding for the corresponding token.
}
  \label{fig:method}

\end{figure*}

\section{\ours~}
\label{sec:approach}

We propose \ours, a relative positional encoding that satisfies all four desiderata introduced previously for multi-view attention. We begin by defining a ray-based position representation (Sec.~\ref{sec:ours-rays}) and formulate relative encoding via projection onto the query frame (Sec.~\ref{sec:ours-rel}). We then introduce expected RoPE encoding to address depth ambiguity (Sec.~\ref{sec:ours-uncertainty}) and finally show that \ours{} can be implemented efficiently, with runtime comparable to baselines (Sec.~\ref{sec:ours-efficiency}).

\subsection{Image Patch as Ray Segments}
\label{sec:ours-rays}
A common way to represent a patch position in 3D is to use the ray starting from the camera center and passing through the patch center, parameterized by the camera center $\mathbf{c}$ and ray direction $\mathbf{r}$. This representation inherently satisfies the uniqueness property. However, it cannot adapt to the geometry of the observed scene, as it is unaware of the depth to the 3D point being intersected by the ray.

% To address this limitation, we generalize the above representation to $(\mathbf{c}, \mathbf{p}^d)$, where $\mathbf{p}^d$ is the point at depth $d$ along the ray. Under homogeneous coordinates, $\mathbf{r}$ is equivalent to $\mathbf{p^\infty}$. To better capture the (typically unknown) 3D scene geometry, we make the model estimate the depth $d$ of the 3D point intersected by the ray, and define our position representation as a ray segment $\mathbf{x}=(\mathbf{c}, \mathbf{p}^d)$ in homogeneous coordinates in the global frame (see Fig.~\ref{fig:method}~(a)).

To address this limitation, we define our position representation as a ray segment in homogeneous coordinates $\mathbf{x}=(\mathbf{c}, \mathbf{p}^d)$, where $\mathbf{p}^d \in \mathbb{P}^3$ is the point where the ray intersects the 3D scene at depth $d$. The ray direction $\mathbf{r}$ is equivalent to $\mathbf{p}^\infty$ (via setting homogeneous term to 0). When a depth map is provided as input, the ray-segment representation can naturally incorporate this information (see Sec.~\ref{sec:broader-app}). In the more common cases where scene geometry is unknown, we make the model estimate the depth $d$ of the 3D point intersected by the ray (see Fig.~\ref{fig:method}~(a)). Specifically, we add a single linear layer to each attention layer to predict the depth $d$. The depth is estimated per-layer; each attention layer now projects input features $\{\tau_i\}$ to compute a depth for each token, which is used to compute the ray encodings. These layers are jointly learned with the model \emph{without any additional supervision}. 
% In practice, we encode each patch with 3 rays passing through 3 corners of the patch instead of 1 ray to fully encode the patch orientation, but for conciseness we assume one ray per patch in the subsequent discussion.

% \ours~naturally extends to multiview transformers that take known depth maps for certain views as inputs: for example, novel-view synthesis (NVS) where the input images and depths are given. Let $d^{\text{known}}$ be the known depth at the pixel intersecting with the ray being encoded. For tokens with known depth available, we can simply replace the predicted depths $d$ with $d^{\text{known}}$ and set the uncertainty $\sigma$ to $0$, while continue using predicted depth for views without known depth (such as target views in NVS). In comparison, previous camera-based RPE methods cannot easily incorporate depth information at the attention level.

\begin{figure}[t]
  \centering
  \includegraphics[width=\columnwidth]{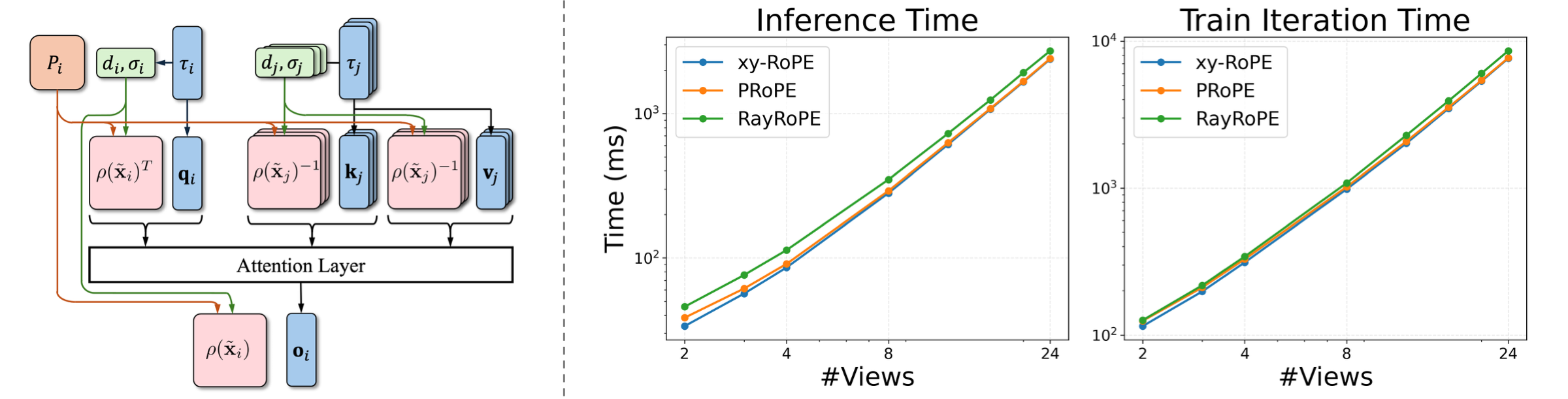}
  \caption{\textbf{Left: Applying \ours~to attention.} We predict per-token depths and uncertainties via a linear projection from features $\mathbf{\tau}$, which are used to compute ray segments. We project the rays with query camera $P_i$, and the computed encodings are applied to $\mathbf{q}, \mathbf{k}, \mathbf{v}$ and $\mathbf{o}$ features.
  \textbf{Right: Efficiency Analysis.} We compare the runtime of different methods. While \ours \ computes additional encodings and depths, this only results in a marginal overhead (13\% at inference and 4\% at training relative to PRoPE for N=3, with a similar trend for larger N).
  }
  \label{fig:apply-attn}
  \vspace{-3mm}
\end{figure}

\subsection{Relative Encodings in Query Frame}
\label{sec:ours-rel}
The ray representation defined above is in the global coordinate frame. To ensure $SE(3)$ invariance, we compute \ours~encodings in each query token's local frame, as illustrated in Fig.~\ref{fig:method}(b). We first define a projective operator that projects global rays into the query camera frame, and then apply standard RoPE to the projected rays in the query frame to support both invariance and multi-frequency similarity.

\noindent\textbf{Projection onto Query Camera.} 
% Given the query camera matrix $P_i=K_iT_i=K_i[R_i|\mathbf{t}_i] \in \mathbb{R}^{3\times 4}$ and an arbitrary ray $(\mathbf{c}, \mathbf{p}^d)$, we define the projected ray as:
% \begin{equation}
%     \tilde{\mathbf{x}} = \pi(P_i, \mathbf{x}) = (T_i\mathbf{c}, P_i\mathbf{p}^d)
%     \label{eq:ours-proj}
% \end{equation}
% where $T_i\mathbf{c}=[x,y,z]^\intercal$ is the 3D camera center transformed into the query frame. $P_i\mathbf{p}^d=[ud',vd',d']^\intercal$ is the projection of the 3D point $\mathbf{p}^d$ onto query camera. We represent it as pixel coordinate $(u,v)$ and disparity $1/d'$. In this way, we can compactly represent a projected ray $\tilde{\mathbf{x}}$ as a 6D vector.
Given the query camera matrix $P_i=K_iT_i \in \mathbb{R}^{4\times 4}$ (assuming $K_i, T_i$ are lifted to $4\times 4$) and an arbitrary ray $(\mathbf{c}, \mathbf{p}^d)$, we define the projected ray as:
\begin{equation}
    \tilde{\mathbf{x}} = \pi(P_i, \mathbf{x}) = (T_i\mathbf{c}, P_i\mathbf{p}^d)
    \label{eq:ours-proj}
\end{equation}
$T_i\mathbf{c}=[c_x,c_y,c_z,1]^\intercal$ represents the 3D camera center transformed into the query frame. $P_i\mathbf{p}^d=[u,v,1,1/d']^\intercal$ is the projection of the 3D point $\mathbf{p}^d$ onto query camera, where $(u,v)$ is the projected pixel coordinate and $1/d'$ is the disparity. Given the transformed (and normalized) homogeneous coordinates, we compactly represent each projected ray $\tilde{\mathbf{x}}$ as a 6D vector: $[c_x, c_y, c_z, u, v, 1/d']$, which will be used to compute the RoPE encodings.

% where $T_i\mathbf{c}=[x,y,z]^\intercal$ is the 3D camera center transformed into the query frame. $P_i\mathbf{p}^d=[ud',vd',d']^\intercal$ is the projection of the 3D point $\mathbf{p}^d$ onto query camera. We represent it as pixel coordinate $(u,v)$ and disparity $1/d'$. In this way, we can compactly represent a projected ray $\tilde{\mathbf{x}}$ as a 6D vector.

\noindent\textbf{Encodings.}
Given the projected rays, we apply RoPE (Eq.~\ref{eq:nd-rope}) with multi-frequency similarity. For a fixed query camera $P_i$, we define the \ours~encoding as $\rho_D(\tilde{\mathbf{x}}) = \rho_D(\pi(P_i, \mathbf{x}))$. Similar to GTA, we apply our encodings to query, key, value, and output features (see Fig.~\ref{fig:apply-attn}):
\begin{equation}
\begin{aligned}
&\text{Attn}^{\text{ours}}(\mathbf{q}_i, \{\mathbf{k}_j,\mathbf{v}_j\}) =\\
\quad \rho_D(\tilde{\mathbf{x}}_i)&\text{Attn}(\rho_D(\tilde{\mathbf{x}}_i)^\intercal \mathbf{q}_i, 
    \{\rho_D(\tilde{\mathbf{x}}_j)^{-1}\mathbf{k}_j, 
    \rho_D(\tilde{\mathbf{x}}_j)^{-1}\mathbf{v}_j\})
\end{aligned}
\label{eq:ours-apply}
\end{equation}
We can show that the above attention expands to the form: 
\begin{equation}
   \frac{\sum_j\exp(q_i^\intercal\rho_D(\tilde{\mathbf{x}}_i-\tilde{\mathbf{x}}_j)k_j)~\rho_D(\tilde{\mathbf{x}}_i- \tilde{\mathbf{x}}_j)v_j}{\sum_j\exp(q_i^\intercal\rho_D(\tilde{\mathbf{x}}_i-\tilde{\mathbf{x}}_j)k_j)}
\label{eq:ours-attn}
\end{equation}
which depends only on the relative positions $\tilde{\mathbf{x}}_i-\tilde{\mathbf{x}}_j$ in the query frame, ensuring $SE(3)$ invariance.

\subsection{Modeling Uncertainty via Expected RoPE}
\label{sec:ours-uncertainty}
\ours~position representation relies on the predicted depth along the rays (Sec.~\ref{sec:ours-rays}). This prediction is only an approximation, and can lead to noisy RoPE encoding, especially for the components where the frequency $\omega$ is high. To alleviate this issue, we predict an uncertainty value $\sigma$ along with each depth $d$. 
As shown in Fig.~\ref{fig:method} (c), this yields a ray segment bounded by two points: $\mathbf{x}^{\text{min}}=(\mathbf{c}, \mathbf{p}^{d-\sigma}),~\mathbf{x}^{\text{max}}=(\mathbf{c}, \mathbf{p}^{d+\sigma})$. Instead of using RoPE on a single position, we propose an expected RoPE encoding $\mathbb{E}_{\tilde{\mathbf{x}}}[\rho_D(\tilde{\mathbf{x}})]$, over the distribution of $\tilde{\mathbf{x}}$.
% assuming projected position $\tilde{\mathbf{x}}$ follows a uniform distribution ranging from $\tilde{\mathbf{x}}^{\text{min}}$ to $\tilde{\mathbf{x}}^{\text{max}}$:
\begin{equation}
    \mathbb{E}_{\tilde{\mathbf{x}}}[\rho_D(\tilde{\mathbf{x}})] = \oplus_{f=1}^{D/2C} \oplus_{c=1}^{C} \mathbb{E}_{x_c}[e^{i\omega_fx_c}]
    \label{eq:ours-expected-rope}
\end{equation}
We assume the projected position $\tilde{\mathbf{x}}$ follows a uniform distribution ranging from $\tilde{\mathbf{x}}^{\text{min}}$ to $\tilde{\mathbf{x}}^{\text{max}}$, i.e., for each component of the ray, $x_c\sim U(x^{\text{min}}, x^{\text{max}})$. We can analytically compute its expected RoPE encoding using the equation below (see Appendix Sec. \ref{supp:implement-expected} for details):
\begin{equation}
\begin{aligned}
    \mathbb{E}_{x_c}
    [e^{i\omega x_c}]
    &= \frac{\int_{x^\text{min}}^{x^{\text{max}}} e^{i\omega x_c}dx_c}{x^{\text{max}}-x^{\text{min}}} = \frac{e^{i\omega x^{\text{max}}}-e^{i\omega x^{\text{min}}}}{i\omega(x^{\text{max}}-x^{\text{min}})}
\end{aligned}    
\label{eq:expected-rope-uniform}
\end{equation}
For deterministic components (e.g., camera centers) where $x^{\text{min}}=x^{\text{max}}$, the expectation reduces to regular RoPE. For positions with larger uncertainty, the expected rotation is `smoothed out' as $x^{\text{max}}-x^{\text{min}}$ increases. The expected RoPE naturally maintains the relative positions when multiplied in attention (assuming positions are independent):
\begin{equation}
    \mathbb{E}_{\tilde{\mathbf{x}}_i}[\rho_D(\tilde{\mathbf{x}}_i)]\mathbb{E}_{\tilde{\mathbf{x}}_j}[\rho_D(\tilde{\mathbf{x}}_j)]^{-1}
    = \mathbb{E}_{\tilde{\mathbf{x}}_i,\tilde{\mathbf{x}}_j}[\rho_D(\tilde{\mathbf{x}}_i-\tilde{\mathbf{x}}_j)]
    \label{eq:expected-rope-relative}
\end{equation}
By modeling uncertainty with expected RoPE, we can produce stable yet geometrically-aware encodings even when predicted depths are approximate.

% \subsection{Extension to Known Depths}
% \label{sec:ours-rgbd}
% \ours~naturally extends to multiview transformers that take known depth maps for certain views as inputs: for example, novel-view synthesis (NVS) where the input images and depths are given. Let $d^{\text{known}}$ be the known depth at the pixel intersecting with the ray being encoded. For tokens with known depth available, we can simply replace the predicted depths $d$ with $d^{\text{known}}$ and set the uncertainty $\sigma$ to $0$, while continue using predicted depth for views without known depth (such as target views in NVS). In comparison, previous camera-based RPE methods cannot easily incorporate depth information at the attention level.

\subsection{Applying \ours \ Efficiently}
\label{sec:ours-efficiency}
While the proposed encodings depends on the query camera, we can efficiently implement it via grouping the attention computation by query view. For each query view, we first compute the corresponding encodings, then perform attention between the query tokens in the current view and all key/value tokens. Because in modern transformers, attention is almost always bottlenecked by the total number of tokens—which remains unchanged—this grouping adds negligible overhead (see Appendix Sec. \ref{supp:implement}).

We benchmark the runtime of \ours \ against regular 2D RoPE \cite{2drope} and PRoPE \cite{prope}. We use the same LVSM setting as in Sec. \ref{sec:exp-nvs}, and summarize results in Fig. \ref{fig:apply-attn}. Despite the query-dependent encodings and depth prediction, \ours's efficiency remains \textit{highly comparable} to the baseline methods, across various number of views $N$. For example, with $N=3$ (which is the default setting in Sec. \ref{sec:exp-nvs}), \ours{} results in a marginal 4\% overhead at training and 13\% at inference when compared to PRoPE. Notably, \ours \ is also compatible with acceleration methods such as KV-caching \cite{pope2023efficiently-kvcache} and FlashAttention \cite{flashattention}.

%% file: sec/5_experiments_reorg.tex
\section{Experiments}
\label{sec:experiments}

We first verify the effectiveness of \ours{} on novel-view synthesis (NVS),
benchmarking it against state-of-the-art positional encodings within the
LVSM~\cite{lvsm} framework (Sec.~\ref{sec:exp-nvs}). We
then ablate the design choices of \ours{} and analyze its internal behavior
(Sec.~\ref{sec:exp-analysis}). Finally, we show that \ours{} extends well
beyond this single setting, improving a diverse range of 3D vision tasks and
model architectures (Sec.~\ref{sec:broader-app}): stereo depth estimation,
feed-forward 3DGS reconstruction, multi-view diffusion.

\begin{table*}[t]
\centering
\scriptsize
\resizebox{1.0\textwidth}{!}{%
\begin{tabular}{l|ccc| ccc| ccc}
\toprule
\multirow{2}{*}{\raisebox{-0.8ex}{Method}}
  & \multicolumn{3}{c|}{CO3D~\cite{co3d}}
  & \multicolumn{3}{c|}{Objaverse~\cite{objaverse}}
  & \multicolumn{3}{c}{RE10K~\cite{re10k}} \\
\cmidrule(lr){2-10}
 & PSNR$\uparrow$ & LPIPS$\downarrow$ & SSIM$\uparrow$
 & PSNR$\uparrow$ & LPIPS$\downarrow$ & SSIM$\uparrow$
 & PSNR$\uparrow$ & LPIPS$\downarrow$ & SSIM$\uparrow$ \\
\midrule
Plucker raymap~\cite{lvsm}
 &16.54 &0.673 &0.538 
 &19.98 &0.273 &0.844 
 &23.45 &0.141 &0.747 \\
RoPE on rays 
 &16.85 &0.589 &0.549 
 &22.17 &0.117 &0.902 
 &25.29 &0.095 &0.809 \\
GTA~\cite{gta}            
 &16.50 &0.602 &0.544 
 &21.87 &0.134 &0.891 
 &24.31 &0.124 &0.774 \\
PRoPE~\cite{prope}          
 &17.49 &0.539 &0.563 
 &22.16 &0.123 &0.896 
 &24.48 &0.112 &0.784 \\
\ours        
 &\textbf{18.40} &\textbf{0.461} &\textbf{0.592}
 &\textbf{22.42} &\textbf{0.110} &\textbf{0.905}
 &\textbf{26.07} &\textbf{0.085} &\textbf{0.831} \\
\midrule
PRoPE (Large)       
 &19.77 &0.329 &0.631 
 &24.66 &0.067 &0.925 
 &27.77 &0.059 &0.866 \\
\ours (Large)     
 &\textbf{20.23} &\textbf{0.308} &\textbf{0.662}
 &\textbf{24.83} &\textbf{0.064} &\textbf{0.929}
 &\textbf{28.31} &\textbf{0.055} &\textbf{0.876} \\
\bottomrule
\end{tabular}
}
\caption{\textbf{Comparison on novel-view synthesis}. When incorporated in LVSM~\cite{lvsm}, \ours \ outperforms all baseline positional encoding methods across three datasets. The bottom two rows show results from models trained at increased scale.}
\label{tab:nvs-results}
\vspace{-5mm}
\end{table*}

\subsection{Novel-View Synthesis}
\label{sec:exp-nvs}

\noindent\textbf{Setup.}
We verify \ours~on novel-view synthesis (NVS) by applying it to LVSM~\cite{lvsm}, a state-of-the-art view synthesis model. Each model takes two posed reference views and the target camera pose, and performs self-attention across all views. Following PRoPE~\cite{prope}, we train downsized LVSM variants ($\sim$47M parameters) from scratch with different positional encoding methods. To demonstrate scalability, we also train larger ($\sim$150M parameters) variants with an $8\times$ batch size. We train models separately on three datasets: CO3D~\cite{co3d}, Objaverse~\cite{objaverse}, and RealEstate10K (RE10K)~\cite{re10k}, and measure reconstruction quality against ground-truth views. For Objaverse, we render a high-quality 80K subset~\cite{lgm} with diverse intrinsics. Except for the large-scale experiments, we train all models with three random seeds and report mean results.

\noindent\textbf{Baselines.}
We compare \ours~against several baseline positional encoding methods. Plucker raymap is the original implementation in LVSM, which concatenates 6D Plucker raymaps to the input tokens. We include the best existing camera-based relative positional encodings: GTA~\cite{gta} and PRoPE~\cite{prope}, both of which are $SE(3)$ invariant. We also design a RoPE-on-rays baseline, which naively applies RoPE to the raymap $(\mathbf{c}, \mathbf{r})$ in global coordinates (as discussed in Sec.~\ref{sec:rope}). Following~\cite{prope}, we concatenate the ``CamRay'' intrinsics raymap to the input for both PRoPE and \ours.

\noindent\textbf{Results.}
As shown in Table~\ref{tab:nvs-results}, \ours~consistently outperforms all baselines across the three datasets at both small and large scales. Overall, relative encodings (GTA, PRoPE, \ours) tend to outperform the absolute Plucker raymap. While it works well on RE10K and Objaverse, the naive RoPE-on-rays baseline struggles on the more challenging CO3D. This highlights the importance of $SE(3)$ invariance for reasoning about complex geometric relationships.

\begin{figure*}[t]
  \centering

  \includegraphics[width=0.99\textwidth]{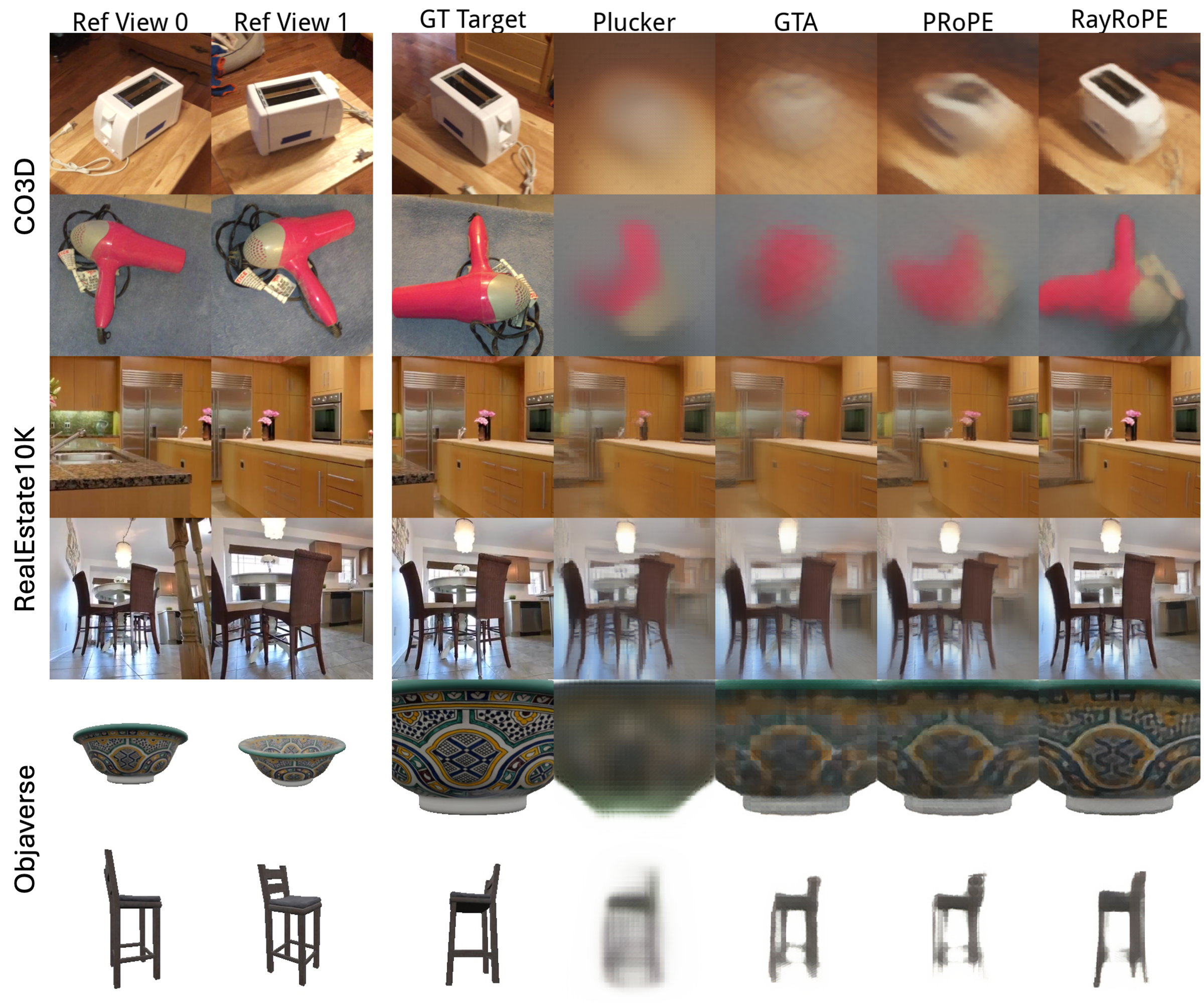}

  \caption{\textbf{Qualitative examples on novel-view synthesis}. \ours~synthesizes more 3D-consistent views with sharper details.}
  \label{fig:nvs-example}
  \vspace{-3mm}
\end{figure*}

In Fig.~\ref{fig:nvs-example}, we visualize several example scenes from each evaluation dataset. For target views that overlap significantly with the reference views, \ours~generates superior high-frequency details compared to baselines. For more challenging target views with little overlap with the reference views, while all methods exhibit a reduction in sharpness, \ours~is nonetheless able to generate more coherent novel views.

\subsection{Ablations and Analysis} \label{sec:exp-analysis}
In this section, we first ablate the key design choices of \ours{}, verifying the importance of the four desiderata in Fig.~\ref{fig:teasor}. We then analyze \ours{}'s behavior by focusing on three different aspects: emergent depth and uncertainty, robustness to out-of-distribution inputs, and the ability to adapt to known depths.

\begin{table}[t]
\centering
\scriptsize
\resizebox{0.99\columnwidth}{!}{
\begin{tabular}{l|ccc|ccc|ccc}
\toprule
\multirow{2}{*}{Method} & \multicolumn{3}{c|}{CO3D~\cite{co3d}} & \multicolumn{3}{c|}{Objaverse~\cite{objaverse}} & \multicolumn{3}{c}{RE10K~\cite{re10k}} \\
\cmidrule(lr){2-4} \cmidrule(lr){5-7} \cmidrule(lr){8-10}
& PSNR$\uparrow$ & LPIPS$\downarrow$ & SSIM$\uparrow$ & PSNR$\uparrow$ & LPIPS$\downarrow$ & SSIM$\uparrow$ & PSNR$\uparrow$ & LPIPS$\downarrow$ & SSIM$\uparrow$ \\
\midrule
\ours~                                        & 18.40 & 0.461 & 0.592 & 22.42 & 0.110 & 0.905 & 26.07 & 0.085 & 0.831 \\
\ding{172} w/o uncertainties            & 17.28 & 0.594 & 0.560 & 20.37 & 0.175 & 0.885 & 25.84 & 0.088 & 0.826 \\
\ding{173} w/o geo-adaptiveness & 17.06 & 0.553 & 0.557 & 22.42 & 0.111 & 0.904 & 25.36 & 0.093 & 0.812 \\
% \ding{174} Use single ray                     & 18.43 & 0.459 & 0.592 & 22.44 & 0.112 & 0.906 & 26.06 & 0.085 & 0.832 \\
\ding{174} w/o multi-frequency                      & 17.49 & 0.550 & 0.563 & 21.37 & 0.223 & 0.862 & 24.27 & 0.115 & 0.779 \\
\ding{175} w/o $\mathbf{v}, \mathbf{o}$ encoding & 18.00 & 0.510 & 0.577 & 21.72 & 0.127 & 0.898 & 25.56 & 0.091 & 0.819 \\
\bottomrule
\end{tabular}
}
\caption{\textbf{Ablations on \ours}. The performance drops of \textbf{Variant} \ding{172}, \ding{173}, \ding{174} highlights the importance of uncertainty modeling, geometric adaptiveness, and multi-frequency similarities, respectively.}
\label{tab:nvs-ablation}
\vspace{-5mm}
\end{table}

\begin{figure*}
  \centering
  \includegraphics[width=\textwidth]{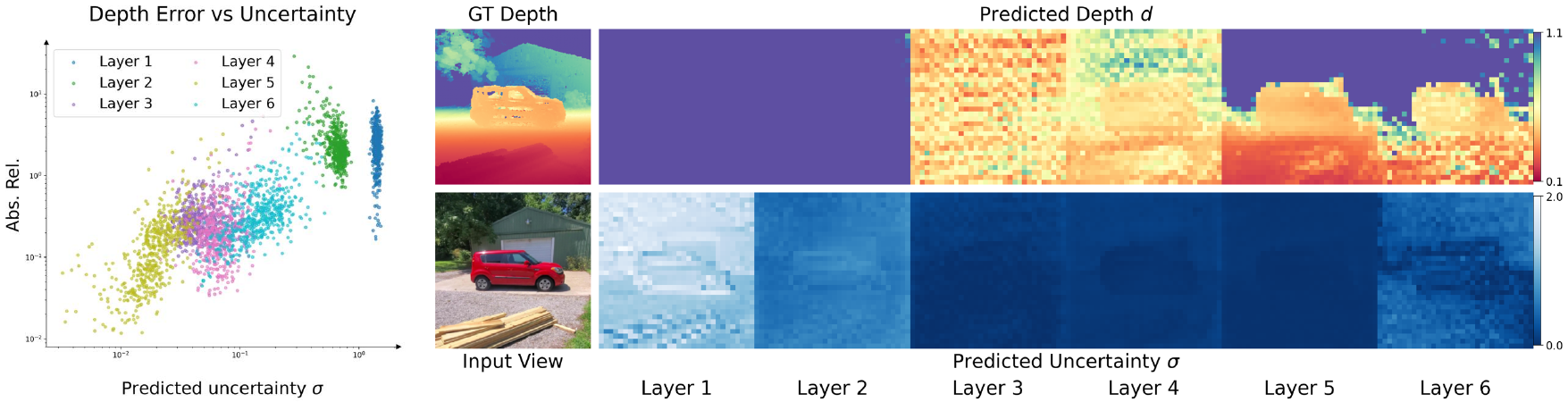}
  \vspace{-3mm}
  \caption{
  \textbf{\textit{Left}: Error on predicted depths vs predicted uncertainties}. In deeper layers (Layers~5--6), the depth errors and uncertainties are strongly positively correlated, showing that the model predicts higher uncertainty when confidence is low. \textbf{\textit{Right}: Predicted depths and uncertainties across layers}. The predicted depth at deeper layers aligns well with the ground-truth depth. The predicted $\sigma$ gradually decreases across layers, showing that confidence in depth prediction gradually improves.
  }
  \label{fig:predicted_d}
  \vspace{-2mm}
\end{figure*}

\noindent\textbf{Ablations}.
In Table~\ref{tab:nvs-ablation}, we ablate the design choices of \ours based on the LVSM experiments in Sec.~\ref{sec:exp-nvs}. Keeping all other configurations the same, we compare the full \ours~to several variants.
\textbf{Variant}~\ding{172}: Remove the uncertainty prediction $\sigma$ (thus disabling expected RoPE).
\textbf{Variant}~\ding{173}: Remove the depth prediction $d$ (thus disabling geometric adaptiveness), using ray direction $\mathbf{p}^\infty$ instead of $\mathbf{p}^d$.
% \textbf{Variant}~\ding{174}: We use only a single ray instead of three rays per patch.
\textbf{Variant}~\ding{174}: Remove the multi-frequency encoding.
\textbf{Variant}~\ding{175}: Remove the encoding on the value and output features ($\mathbf{v}, \mathbf{o}$), applying it only to query and key features.

All four variants degrade performance to varying extents. Variant~\ding{172} shows that uncertainty modeling via expected RoPE is crucial for robust performance, especially when pose variations are larger (CO3D and Objaverse). Variant~\ding{173} highlights the importance of geometric adaptiveness via depth prediction, while Variant~\ding{174} highlights the benefits of multi-frequency encodings. Variant~\ding{175} shows that applying the encodings to value and output features is helpful, consistent with prior work~\cite{gta}.

\noindent\textbf{Analysis of Emergent Depth.}
We analyze the predicted depth $d$ (Sec.~\ref{sec:ours-rays}) and uncertainty $\sigma$ (Sec.~\ref{sec:ours-uncertainty}) and show that they encode meaningful geometric information. We measure the error of predicted depths on CO3D scenes and plot it against the predicted uncertainties (left panel of Fig.~\ref{fig:predicted_d}). Each point corresponds to one image at a specific layer. The model exhibits high uncertainty in the early layers, consistent with ambiguity at the beginning of processing. As features propagate through the network (Layers~5--6), the uncertainty decreases and shows a strong positive correlation with depth error, indicating that higher uncertainty is assigned to less reliable predictions. \ours~subsequently incorporates such uncertainty by computing the expected RoPE. The right panel of Fig.~\ref{fig:predicted_d} visualizes the evolution of the predicted depth and uncertainty maps across all six layers on a representative scene. By Layer~5, geometrically plausible depth maps emerge despite the absence of depth supervision. This result supports the conclusion that \ours~leverages depth predictions in a geometrically meaningful manner. See Appendix Fig. \ref{fig:extra-pred-depths} for more visual examples.

\begin{figure}[t]
  \centering
  \includegraphics[width=\columnwidth]{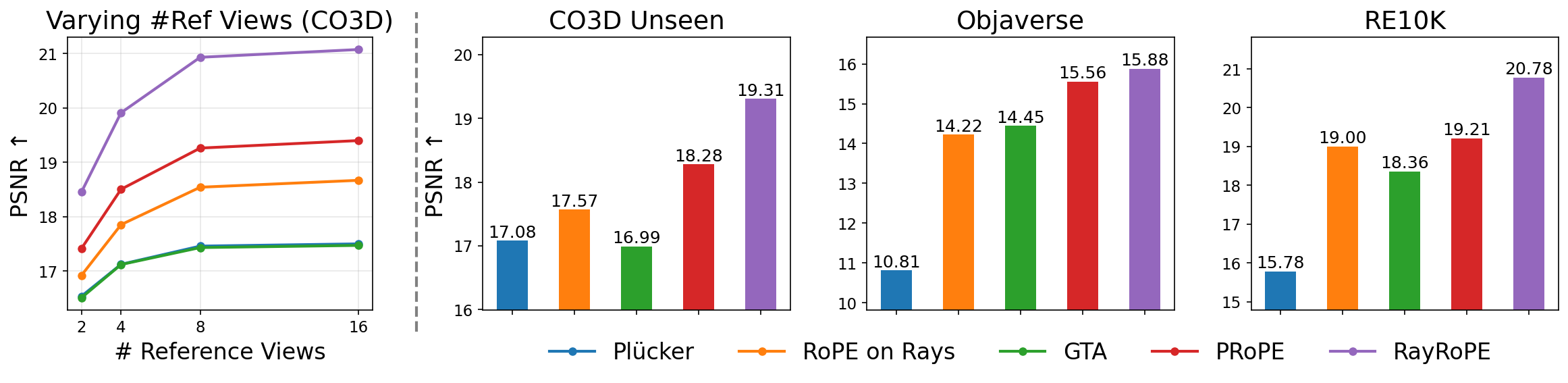}
  \vspace{-6mm}
  \caption{\textbf{Out-of-distribution Robustness.}
  \textbf{\textit{Left}}: Evaluation on varying numbers of reference views (trained only with two reference views). \textbf{\textit{Right}}: Evaluating models trained on CO3D on out-of-distribution categories and datasets. \ours\ maintains its advantage in both settings. Full results are in Appendix Fig. \ref{fig:var-ref-views}}
  \label{fig:generalization-psnr}
  \vspace{-2mm}
\end{figure}

\noindent\textbf{Out-of-distribution Robustness}
To evaluate the robustness of \ours\ on out-of-distribution tasks, we evaluate the LVSM models from the previous experiments in a zero-shot manner under two distinct settings: \textbf{\textit{(a) varying reference views:}} Given the models trained with two reference views, we evaluate inference performance when provided with (2, 4, 8, 16) views.
% The advantage of \ours \ slightly widens as number of reference views increases, showing a strong robustness to number of reference views. While the performance could decline when $\#\textbf{views} \ge 8$ due to out-of distribution settings, our method decline less in most settings, showing a stronger robustness.
\textbf{\textit{(b) out-of-domain data:}} We take the LVSM models trained exclusively on CO3D and evaluate them on unseen domains, including RE10K, Objaverse, and unseen categories of CO3D.

We show representative results in Fig.~\ref{fig:generalization-psnr}, while full results are in Appendix Fig. \ref{fig:var-ref-views}. In both settings, \ours\ maintains its superior performance over previous methods, showing strong robustness to unseen numbers of reference views and datasets. These results also verify that depth prediction does not undermine the robustness of \ours, even in out-of-distribution settings.

\noindent\textbf{Adapting to known depth}.
To further highlight \ours{}'s adaptability to scene geometry, we train and evaluate LVSM with known depths for reference views. As discussed in Sec.~\ref{sec:ours-rays}, \ours{} can seamlessly incorporate known depths by replacing the predicted depth $d$ with the ground-truth value, while existing methods such as PRoPE cannot naturally make use of such information. We set uncertainties to zero when depths are known, and retain the predicted $d, \sigma$ for target views. For all models, we concatenate the depth maps of the reference views at the input. As shown in Table~\ref{tab:lvsm-depth}, the depth-aware \ours{} yields larger gains over its RGB-only counterpart, highlighting the benefit of geometry-aware positional encoding in multi-view attention.

\begin{table}[t]
\centering
\scriptsize
\resizebox{0.7\columnwidth}{!}{
\begin{tabular}{l|c|ccc|ccc}
\toprule
\multirow{2}{*}{Method} & \multirow{2}{*}{Depth} & \multicolumn{3}{c|}{CO3D~\cite{co3d}} & \multicolumn{3}{c}{Objaverse~\cite{objaverse}} \\
\cmidrule(lr){3-5} \cmidrule(lr){6-8}
& & PSNR$\uparrow$ & LPIPS$\downarrow$ & SSIM$\uparrow$ & PSNR$\uparrow$ & LPIPS$\downarrow$ & SSIM$\uparrow$ \\ 
\midrule
\multirow{2}{*}{PRoPE~\cite{prope}} & \xmark & 17.49 & 0.539 & 0.563 & 22.16 & 0.123 & 0.896 \\
                             & \cmark & 19.10 & 0.434 & 0.611 & 23.20 & 0.110 & 0.903 \\
\multirow{2}{*}{\ours}       & \xmark & 18.40 & 0.461 & 0.592 & 22.42 & 0.110 & 0.905 \\
                             & \cmark & \textbf{20.47} & \textbf{0.284} & \textbf{0.692} & \textbf{25.19} & \textbf{0.067} & \textbf{0.929} \\
\bottomrule
\end{tabular}
}
\caption{\textbf{NVS with known depth}. We train LVSM with known ground-truth depths concatenated to the reference views. \ours\ can incorporate this information by replacing the predicted depths with the known ones, yielding larger improvements compared to its counterpart trained without depths.}
\label{tab:lvsm-depth}
\vspace{-5mm}
\end{table}

\subsection{Broader Applications}
\label{sec:broader-app}
We have evaluated and analyzed \ours{} on the LVSM framework for novel-view synthesis. In this section, we show that \ours{} can serve as a plug-and-play mechanism that is generally applicable to any multi-view attention module with camera conditioning. Across stereo depth estimation \cite{unimatch}, feed-forward 3DGS reconstruction \cite{tokengs}, and multi-view diffusion \cite{cape}, \ours{} consistently improves over the respective baselines.

\begin{figure*}[t]
  \centering

  \includegraphics[width=0.99\textwidth]{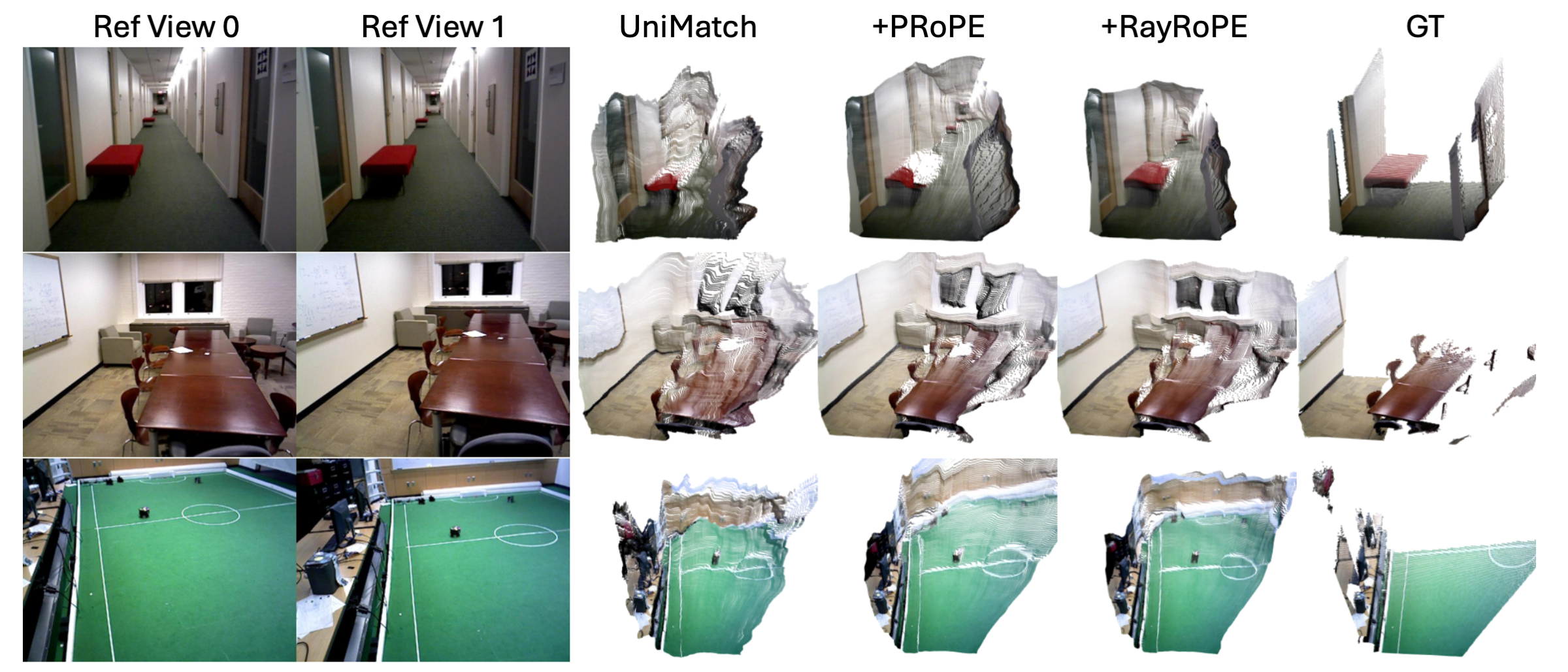}
  \vspace{-3mm}
  \caption{\textbf{Example 3D points reprojected from stereo depth estimation.} \ours~produces more accurate depth predictions and thus better 3D reconstruction.}
  \label{fig:stereo}
\end{figure*}

\noindent\textbf{Stereo Depth Estimation}
We adopt UniMatch~\cite{unimatch}, a multi-view transformer trained for multiple tasks, including stereo depth estimation. The model takes stereo image pairs with known poses. We apply \ours\ to the cross-image attention layers of UniMatch. Following the original work, we train and evaluate the model on RGBD~\cite{rgbd}, SUN3D~\cite{sun3d}, and Scenes11~\cite{DeMon}. To evaluate out-of-distribution performance, we also benchmark on an unseen dataset, ScanNet~\cite{scannet}. As shown in Tab.~\ref{tab:stereo_results}, compared to both the original UniMatch and the variant with PRoPE, the model with \ours\ achieves more accurate depth predictions. Qualitative results in Fig.~\ref{fig:stereo} further illustrate improved geometric consistency, showing that \ours\ yields more accurate depth structures.

\begin{table}[t]
\centering
\setlength{\tabcolsep}{3pt}
\renewcommand{\arraystretch}{1.2}
\scriptsize
\resizebox{\columnwidth}{!}{
\begin{tabular}{l| cc|cc| cc|cc}
\toprule
\multirow{2}{*}{Method} & \multicolumn{2}{c|}{RGBD~\cite{rgbd}} & \multicolumn{2}{c|}{SUN3D~\cite{sun3d}} & \multicolumn{2}{c|}{Scenes11~\cite{DeMon}} & \multicolumn{2}{c}{ScanNet~\cite{scannet}(Unseen)} \\
\cmidrule(lr){2-3} \cmidrule(lr){4-5} \cmidrule(lr){6-7} \cmidrule(lr){8-9}
& Abs Rel$\downarrow$ & RMSE$\downarrow$ & Abs Rel$\downarrow$ & RMSE$\downarrow$ & Abs Rel$\downarrow$ & RMSE$\downarrow$ & Abs Rel$\downarrow$ & RMSE$\downarrow$ \\
\midrule
UniMatch~\cite{unimatch} & 0.119 & 0.628 & 0.135 & 0.406 & 0.086 & 0.742 & 0.127 & 0.346\\
+PRoPE~\cite{prope}      & \textbf{0.106} & 0.589 & 0.113 &  0.338 & 0.051 & 0.494 & 0.101 &  0.285 \\
+\ours                  & \textbf{0.106} & \textbf{0.574} &  \textbf{0.109} & \textbf{0.328} & \textbf{0.047} & \textbf{0.473} & \textbf{0.095} & \textbf{0.276}  \\
\bottomrule
\end{tabular}
}
\caption{\textbf{Comparisons on stereo depth estimation.} We report absolute relative error (Abs. Rel.) and root mean square error (RMSE) across four datasets, including the out-of-distribution ScanNet~\cite{scannet}. The UniMatch~\cite{unimatch} model with \ours{} encoding predicts more accurate depths than the baselines.}
\label{tab:stereo_results}
\vspace{-5mm}
\end{table}

\begin{table}[t]
\centering
% \scriptsize
\tiny
\resizebox{\columnwidth}{!}{%
\begin{tabular}{l|ccc|ccc|ccc}
\toprule
\multirow{2}{*}{\raisebox{-0.8ex}{Method}}
  & \multicolumn{3}{c|}{2 views}
  & \multicolumn{3}{c|}{4 views}
  & \multicolumn{3}{c}{6 views} \\
\cmidrule(lr){2-4}\cmidrule(lr){5-7}\cmidrule(lr){8-10}
 & PSNR$\uparrow$ & LPIPS$\downarrow$ & SSIM$\uparrow$
 & PSNR$\uparrow$ & LPIPS$\downarrow$ & SSIM$\uparrow$
 & PSNR$\uparrow$ & LPIPS$\downarrow$ & SSIM$\uparrow$ \\
\midrule
TokenGS~\cite{tokengs}
 & 18.84 & 0.467 & 0.584 & 22.85 & 0.363 & 0.732 & 23.83 & 0.343 & 0.760 \\
+PRoPE~\cite{prope}
 & 19.02 & 0.461 & \textbf{0.598} & 23.04 & 0.361 & 0.741 & 24.09 & 0.342 & 0.770 \\
+\ours
 & \textbf{19.15} & \textbf{0.457} & \textbf{0.598} & \textbf{23.40} & \textbf{0.339} & \textbf{0.758} & \textbf{24.63} & \textbf{0.312} & \textbf{0.793} \\
\bottomrule
\end{tabular}
}
\caption{\textbf{Comparison on feed-forward 3DGS reconstruction.} We adapt TokenGS \cite{tokengs}, a feed-forward model that predict 3DGS from posed input views. We benchmark the rendering quality of the output 3D Gaussian splats given 2, 4, and 6 input views. \ours{} helps the model to predict better Gaussian splats.}
\label{tab:tokengs}

\vspace{-4mm}
\end{table}

\begin{table}[t]
\centering
% \scriptsize
\tiny
\resizebox{\columnwidth}{!}{%
\begin{tabular}{l|ccc|ccc|ccc}
\toprule
\multirow{2}{*}{\raisebox{-0.8ex}{Method}}
  & \multicolumn{3}{c|}{3 views}
  & \multicolumn{3}{c|}{6 views}
  & \multicolumn{3}{c}{9 views} \\
\cmidrule(lr){2-4}\cmidrule(lr){5-7}\cmidrule(lr){8-10}
 & PSNR$\uparrow$ & LPIPS$\downarrow$ & SSIM$\uparrow$
 & PSNR$\uparrow$ & LPIPS$\downarrow$ & SSIM$\uparrow$
 & PSNR$\uparrow$ & LPIPS$\downarrow$ & SSIM$\uparrow$ \\
\midrule
EscherNet~\cite{cape}            
 &15.81 &0.279 &0.373 &16.31 &0.246 &0.391 &16.42 &0.236 &0.393 \\
+PRoPE~\cite{prope}          
 &16.87 &0.254 &0.434 &17.85 &0.211 &0.480 &18.36 &0.189 &0.504 \\
+\ours        
 &\textbf{17.14} &\textbf{0.239} &\textbf{0.464} &\textbf{18.16} &\textbf{0.209} &\textbf{0.504} &\textbf{19.03} &\textbf{0.188} &\textbf{0.543} \\
\bottomrule
\end{tabular}
}
\caption{\textbf{Comparison on Multi-view diffusion.} We apply \ours{} to EscherNet~\cite{cape}, a large multi-view diffusion model for view synthesis. On DL3DV-10K~\cite{dl3dv} with 3, 6, or 9 reference views, \ours{} consistently outperforms both baselines.}
\label{tab:eschernet}
\vspace{-4mm}

\end{table}

\noindent\textbf{Feed-forward 3D Gaussian Splatting Reconstruction}.
We also experiment with TokenGS~\cite{tokengs}, a recent feed-forward 3DGS~\cite{3dgs} reconstruction model. TokenGS first encodes posed multi-view inputs into multi-view features and then decodes a set of scene-level 3DGS tokens into 3DGS parameters by cross-attending to the multi-view features. We apply \ours{} to the encoder self-attention and the image-to-3DGS cross-attention. In the cross-attention, we apply \ours{} only to the image features and skip ray projection, since the 3DGS query tokens do not have explicit positions. Following the original work's settings, we train on DL3DV-10K~\cite{dl3dv} with four input views and evaluate the rendering quality of the predicted 3DGS. As shown in Table~\ref{tab:tokengs} and Fig.~\ref{fig:example-tokengs}, we find that \ours{} outperforms both the original method and the PRoPE variant. These results suggest that our method generalizes across architectures (e.g., cross-attention between image features and implicit latent tokens).

\noindent\textbf{Multi-view diffusion model}.
We adopt EscherNet~\cite{cape}, a large camera-controlled multi-view diffusion model. We retrain EscherNet variants on DL3DV-10K~\cite{dl3dv} with CaPE (the original design), PRoPE, and \ours{}. As shown in Table~\ref{tab:eschernet} and Fig.~\ref{fig:example-eschernet}, \ours{} remains advantageous, with the performance gap widening as the number of reference views increases. These results also indicate that \ours{}'s depth predictions remain robust even on the noisy images in diffusion models.

% We apply \ours{} to EscherNet~\cite{cape}, a large ($\sim$800M parameters) multi-view conditioned diffusion model for novel view synthesis. Unlike LVSM that applies self-attention on all views, EscherNet performs cross-attention between reference and target views. We replace the original CaPE positional encoding in EscherNet with PRoPE or \ours{}, and train/evaluate the resulting variants on DL3DV~\cite{dl3dv}. Given a varying number of reference views, the model generates three target views, and we evaluate the results using standard metrics. The results are summarized in Table~\ref{tab:eschernet} and illustrated in Fig.~\ref{fig:example-eschernet}. Across different numbers of reference views, \ours{} consistently outperforms both the original model and the PRoPE variant, with the performance gap increasing as the number of reference views grows. These results verifies the effectiveness of \ours \ when scaled to larger model and a greater number of views.

%% file: sec/6_discussion.tex
\section{Discussion}
% We presented \ours, a position encoding for multi-view transformers that process a set of posed input images. While \ours~outperforms prior formulations, there are areas for further improvement and investigation. In particular, while \ours~could incorporate uncertainty in predicting depths along the ray, it would also be interesting to model the uncertainty in the camera matrices. More broadly, while this work focused on positional embeddings for \emph{posed} input images, designing such encodings for multi-view transformers that process unposed (or mixed) images remains an open challenge.

In this work, we identified four desirable properties for positional encodings in multi-view attention (Fig.~\ref{fig:teasor}). Guided by these criteria, we introduced \ours, a ray-based positional encoding that processes a set of posed input images. \ours{} consistently improves performance on various tasks and models in 3D vision, showing that satisfying these properties yields more effective multi-view attention. While \ours{} could incorporate uncertainty when predicting depth along a ray, it would also be interesting to model uncertainty in the camera matrices. More broadly, while this work focused on positional embeddings for \emph{posed} input images, designing such encodings for multi-view transformers that process unposed (or mixed) images remains an open challenge.

\section*{Acknowledgment}
\label{sec:acknowledgement}
We would like to thank Zihan Wang and Qitao Zhao for insightful discussions about the project. This project was supported by Apple and NSF Award IIS2345610. This work used computation resources at NCSA Delta, NCSA Delta-AI, and Bridges-2 through allocation CIS240289, CIS251333, and CIS240132 from the Advanced Cyberinfrastructure Coordination Ecosystem: Services \& Support (ACCESS) program, which is supported by U.S. National Science Foundation grants \#2138259, \#2138286, \#2138307, \#2137603, and \#2138296. 

% \lucas{Camera-Ready Note:\\
% \textbf{Changes made}:\\
% Section 5.4: Renamed from "NVS with known depth" to "Architecture Generalization"\\
% Renamed Section 5.3 \\
% cite new relevant work\\
% mention the new results in abstract/intro\\
% Restructure the Section 5 \\
% Appendix B: experimental details on eschernet and tokengs\\
% refer to appendix in main text\\
% fixed citation arxiv-> conference \\
% \textbf{ToDos}:\\
% }

%% file: sec/X_suppl.tex
\clearpage
\appendix
% ECCV/LLNCS doesn't define \maketitlesupplementary (CVPR macro), so we just start the appendix here.
% \setcounter{page}{1} % uncomment if you want appendix page numbers to restart

% \section{Rationale}
% \label{sec:rationale}
% % 
% Having the supplementary compiled together with the main paper means that:
% % 
% \begin{itemize}
% \item The supplementary can back-reference sections of the main paper, for example, we can refer to \cref{sec:intro};
% \item The main paper can forward reference sub-sections within the supplementary explicitly (e.g. referring to a particular experiment); 
% \item When submitted to arXiv, the supplementary will already included at the end of the paper.
% \end{itemize}
% % 
% To split the supplementary pages from the main paper, you can use \href{https://support.apple.com/en-ca/guide/preview/prvw11793/mac#:~:text=Delete%20a%20page%20from%20a,or%20choose%20Edit%20%3E%20Delete).}{Preview (on macOS)}, \href{https://www.adobe.com/acrobat/how-to/delete-pages-from-pdf.html#:~:text=Choose%20%E2%80%9CTools%E2%80%9D%20%3E%20%E2%80%9COrganize,or%20pages%20from%20the%20file.}{Adobe Acrobat} (on all OSs), as well as \href{https://superuser.com/questions/517986/is-it-possible-to-delete-some-pages-of-a-pdf-document}{command line tools}.

\section{Implementing RayRoPE}
\subsection{Details on Applying \ours}
\label{supp:implement}
As briefly discussed in Sec.~\ref{sec:ours-efficiency}, the \ours~encoding is coupled with the camera coordinates of the query tokens. Consequently, our implementation groups query tokens by their corresponding camera views to compute view-dependent encodings and attention. We detail the multiview self-attention procedure with \ours~in Algorithm~\ref{algo:ours}, which can be straightforwardly generalized to cross-attention.

Given a set of input token features $\tau \in \mathbb{R}^{N \times HW \times D}$ from $N$ views, we first perform linear projections to obtain the standard query, key, and value features. Simultaneously, we project $\tau$ via linear layers $W_d$ and $W_\sigma$ to predict the ray depth $d$ and uncertainty $\sigma$, respectively (we omit bias terms in Algorithm~\ref{algo:ours} for conciseness). Based on $d$, $\sigma$, and the camera poses $P$, we construct the ray representation $\mathbf{x}$ in the global coordinate frame, as defined in Sec.~\ref{sec:ours-rays}.

We then execute attention iteratively for each query camera to account for relative geometric transformations. For a specific query view $n$, we denote the corresponding query tokens and camera pose as $Q[n]$ and $P[n]$, respectively, where $[*]$ denotes indexing into a tensor. Following Eq.~\ref{eq:ours-proj}, we project all ray segments $\mathbf{x}$ into the local coordinates of the $n$-th camera, yielding projected rays $\mathbf{\tilde{x}} = \pi(P[n], \mathbf{x})$. We then compute the \ours~encoding based on the expected RoPE of these projected positions (Eq.~\ref{eq:ours-expected-rope}), obtaining an encoding tensor $\text{Enc} \in \mathbb{R}^{N\times HW \times D \times D}$. We apply the \ours~encodings to the query subset $Q[n]$, all keys and values, and the attention output (see Fig.~\ref{fig:apply-attn}). The final output tensor $O$ is formed by concatenating the processed features $\{O_1, O_2, \dots, O_N\}$ across the view dimension.

\begin{algorithm}[t]
\caption{Multiview Self-Attention with \ours}
\label{algo:ours}
\begin{algorithmic}[1]
    % 1. INPUTS
    % In 'algorithmic', \item[] creates an unnumbered line similar to \Statex
    \REQUIRE
    \item[] Number of views $N$, feature dimension $D$
    \item[] Number of tokens along height, width $(H,W)$
    \item[] Token features $\tau \in \mathbb{R}^{N \times HW \times D}$
    \item[] Linear layer $W_q, W_k, W_v \in \mathbb{R}^{D \times D}$
    \item[] Linear layer $W_d, W_\sigma \in \mathbb{R}^{D \times 1}$
    \item[] Camera poses $P \in \mathbb{R}^{N \times 4 \times 4}$
    % 2. OUTPUTS
    \ENSURE
    \item[] Output features $O \in \mathbb{R}^{N \times HW \times D}$

    \item[] \hrulefill % Divider line
    \STATE $Q,~K,~V \leftarrow W_q\tau,~W_k\tau,~W_v\tau$
    \STATE $d,~\sigma \leftarrow \exp(W_d\tau),~\exp(W_\sigma \tau)$
    \STATE $\mathbf{x} \leftarrow \text{get\_rays}(P,~d,~\sigma )$

    \FOR{$n = 1$ \textbf{to} $N$}
        \STATE $\mathbf{\tilde{x}} \leftarrow \pi(P[n], \mathbf{x})$
        \STATE $\text{Enc} \leftarrow \text{get\_encoding}(\mathbf{\tilde{x}})$
        \STATE $Q_n \leftarrow \text{Enc}^\intercal[n]Q[n]$
        \STATE $K_n \leftarrow \text{Enc}^{-1} K$
        \STATE $V_n \leftarrow \text{Enc}^{-1} V$
        \STATE $O_n \leftarrow \mathrm{Attn}(Q_n, K_n, V_n)$
        \STATE $O_n \leftarrow \text{Enc}[n]O_n$

    \ENDFOR
    \STATE $O \leftarrow \text{concatenate}(\{O_1, O_2, \dots, O_N\})$
    \RETURN $O$
\end{algorithmic}
\end{algorithm}

\subsection{Details on Expected RoPE}
\label{supp:implement-expected}
\noindent\textbf{Implementation.}
As introduced in Sec.~\ref{sec:ours-uncertainty}, \ours\ assumes a uniform distribution for the projected position: $\tilde{\mathbf{x}} \sim U(\tilde{\mathbf{x}}^{\text{min}}, \tilde{\mathbf{x}}^{\text{max}})$. 
For convenience, we restate Eqs.~\ref{eq:ours-expected-rope} and~\ref{eq:expected-rope-uniform} here:
\begin{equation*}
    \mathbb{E}_{\tilde{\mathbf{x}}}[\rho_D(\tilde{\mathbf{x}})] = \oplus_{f=1}^{D/2C} \oplus_{c=1}^{C} \mathbb{E}_{x_c}[e^{i\omega_fx_c}]
    \tag{8}
\end{equation*}
\begin{equation*}
    \quad \mathbb{E}_{x_c} [e^{i\omega x_c}] = \frac{e^{i\omega x^{\text{max}}}-e^{i\omega x^{\text{min}}}}{i\omega(x^{\text{max}}-x^{\text{min}})}
    \tag{9}
\end{equation*}
To concretize this for implementation in standard deep learning frameworks, we write the complex exponential as an $SO(2)$ rotation matrix:
\begin{equation*}
    e^{i \omega x} = 
    \begin{bmatrix} 
    \cos(\omega x) & -\sin(\omega x) \\ 
    \sin(\omega x) & \cos(\omega x) 
    \end{bmatrix}
\end{equation*}
Substituting this into Eq.~\ref{eq:expected-rope-uniform} gives:
% \\begin{equation}
%     \\mathbb{E}_{x}[e^{i \\omega x_c}] = \\frac{1}{\\omega \\Delta x}
%     \\begin{bmatrix} 
%     \\sin(\\omega x^{\\text{max}}) - \\sin(\\omega x^{\\text{min}}) & \\cos(\\omega x^{\\text{max}}) - \\cos(\\omega x^{\\text{min}}) \\\\ 
%     \\cos(\\omega x^{\\text{min}}) - \\cos(\\omega x^{\\text{max}}) & \\sin(\\omega x^{\\text{max}}) - \\sin(\\omega x^{\\text{min}})
%     \\end{bmatrix}
%     \\label{eq:supp-matrix-form}
% \\end{equation}
\begin{equation*}
\begin{split}
&\mathbb{E}_{x}\big[e^{i\omega x_c}\big] =
\frac{1}{\omega (x^{\text{max}}-x^{\text{min}})} \\[4pt]
&\times{
\begin{bmatrix}
\sin(\omega x^{\text{max}})-\sin(\omega x^{\text{min}}) &
\cos(\omega x^{\text{max}})-\cos(\omega x^{\text{min}}) \\
\cos(\omega x^{\text{min}})-\cos(\omega x^{\text{max}}) &
\sin(\omega x^{\text{max}})-\sin(\omega x^{\text{min}})
\end{bmatrix}
}
\end{split}
\end{equation*}
All elements in the above form are composed of standard trigonometric functions and can be computed efficiently.

\noindent\textbf{Proof of Relative Position Preservation.}
We provide a proof of Eq.~\ref{eq:expected-rope-relative}, which states that the expected RoPE depends only on the relative position between two tokens. We assume the projected positions $\tilde{\mathbf{x}}_i$ and $\tilde{\mathbf{x}}_j$ are independent random variables. Using the definition $\rho(x)=e^{i\omega x}$, we expand the expectation of the relative position encoding (the RHS of Eq.~\ref{eq:expected-rope-relative} in the main text) and show that it equals the product of the expected encodings:
\begin{equation}
\begin{aligned}
    \mathbb{E}_{\tilde{\mathbf{x}}_i,\tilde{\mathbf{x}}_j}[\rho_D(\tilde{\mathbf{x}}_i-\tilde{\mathbf{x}}_j)]
    &= \mathbb{E}_{\tilde{\mathbf{x}}_i, \tilde{\mathbf{x}}_j} \left[ e^{i\omega (\tilde{\mathbf{x}}_i - \tilde{\mathbf{x}}_j)} \right] \\
    &= \mathbb{E}_{\tilde{\mathbf{x}}_i, \tilde{\mathbf{x}}_j} \left[ e^{i\omega \tilde{\mathbf{x}}_i} \cdot e^{-i\omega \tilde{\mathbf{x}}_j} \right] \\
    &= \mathbb{E}_{\tilde{\mathbf{x}}_i} [ e^{i\omega \tilde{\mathbf{x}}_i} ] \cdot \mathbb{E}_{\tilde{\mathbf{x}}_j} [ e^{-i\omega \tilde{\mathbf{x}}_j} ] \\
    &= \mathbb{E}_{\tilde{\mathbf{x}}_i} [ \rho(\tilde{\mathbf{x}}_i) ] \cdot \mathbb{E}_{\tilde{\mathbf{x}}_j} [ \rho(\tilde{\mathbf{x}}_j)^{-1} ]
\end{aligned}
\end{equation}
This is equivalent to Eq.~\ref{eq:expected-rope-relative}, confirming that the attention output between two tokens with expected-RoPE embeddings is a function of their relative ray position and is independent of their absolute position.

\section{Experimental Details}
\noindent \textbf{LVSM}. We adapt the LVSM~\cite{lvsm} model by following the codebase released by PRoPE~\cite{prope}. In all of our experiments, we use 6 transformer layers and a feedforward dimension of 1024. We set the attention feature dimension to 1152, distributed over 8 attention heads. This yields a smaller version of LVSM with around 47M parameters. We train on 2 GPUs with a total batch size of 8. For all datasets, we fix the number of reference views to 2 and the number of supervised views to 1. We normalize all input camera poses such that the first view's extrinsics are the identity. We train exclusively at $(256, 256)$ resolution.

Following LVSM, we use the same train/test split for RE10K~\cite{re10k} as in pixelSplat~\cite{pixelsplat}. For Objaverse~\cite{objaverse}, we use a high-quality subset of size 80K filtered by LGM~\cite{lgm}. We randomly sample 10\% of scenes as a held-out validation set. For each scene, we randomly sample 8 sets of elevation and azimuth viewing angles. For each viewing angle, we construct three cameras with a random field of view and a random distance from the world center. For CO3D~\cite{co3d}, we follow the split in~\cite{camera_as_rays}. We train on 41 categories and evaluate on held-out scenes within the same categories. We also evaluate on the 10 held-out categories in Sec.~\ref{sec:add-result}.

To train the larger LVSM model (the bottom two rows in Table~\ref{tab:nvs-results}), we use 12 transformer layers and set the feedforward dimension to 3072. This results in a model with around 150M parameters. We train with a total batch size of 64, distributed over 8 GPUs. All other configurations are kept the same.

\noindent \textbf{Unimatch}. We adapt the official codebase~\cite{unimatch}. We set the attention feature dimension to 144 and train with an effective batch size of 16. All other training and evaluation configurations are kept exactly the same as in the original paper.

\noindent \textbf{TokenGS}. Following the official implementation \cite{tokengs}, we train on the DL3DV-10K~\cite{dl3dv} dataset and evaluate on the DL3DV-10K Benchmark dataset. We set the feature dimension to 1152. All other configurations are kept consistent with the paper: we first train at $(256, 256)$ resolution with 1024 latent 3DGS tokens for 150K iterations, then finetune at $(256, 448)$ resolution with 4096 3DGS tokens for 10K iterations. We use a batch size of 64 and 4 input views during training. For all evaluations, we apply the test-time tuning to the predicted 3DGS.

We apply \ours{} to the multi-view self-attention and the image-to-3DGS cross-attention. Since the 3DGS latent tokens (i.e., query tokens) in cross-attention do not have any explicit positions, we skip the query-frame projection process and only apply ray encoding to the image features in the global coordinate system. Using the global coordinate system is acceptable here because TokenGS predicts 3DGS in the same global coordinate system, making $SE(3)$ invariance inherently infeasible. We apply PRoPE with the same scheme.

\noindent \textbf{EscherNet}. We train and evaluate on a different dataset (DL3DV-10K) from the original work. We use 3 input views and 3 target views during training. We train each model for 70K iterations with an effective batch size of 32.

% \section{Runtime Efficiency}

% Although our method necessitates computing $N$ sets of encodings and group-wise attention (Sec.~\ref{supp:implement}), we empirically find that \ours~exhibits efficiency highly comparable to the baselines. We benchmark the runtime of \ours~against the regular 2D RoPE with patch indices (xy-RoPE~\cite{2drope}) and PRoPE~\cite{prope}. We utilize the identical LVSM architecture employed in the main experiment. We measure the wall-clock time per iteration for both the inference and training phases. All profiling was performed on a single NVIDIA RTX A6000 GPU. The quantitative results are summarized in Fig.~\ref{fig:runtime}.

% In the top row of Fig.~\ref{fig:runtime}, we fix the number of views to 3 and compare the per-iteration runtime. We observe only a marginal runtime overhead (13\% during inference and 4\% during training relative to PRoPE). This suggests that the geometric precision offered by RayRoPE is achieved with marginal cost to system throughput. The bottom row of Fig.~\ref{fig:runtime} demonstrates that RayRoPE maintains computational efficiency as the number of input views increases. We observe that the scaling trend are the same across all methods, indicating that our approach scales effectively alongside the baselines.

% \begin{figure}[t]
%   \centering
%   \includegraphics[width=\columnwidth]{figs/efficiency_summary.png}
%   \caption{ \textbf{Comparison on runtime efficiency}. RayRoPE maintains a runtime efficiency highly comparable to PRoPE, with only a marginal overhead.}
%   \label{fig:runtime}
% \end{figure}

\section{Analysis of Attention Similarity}
\begin{figure*}[t]
  \centering
  \includegraphics[width=0.95\textwidth]{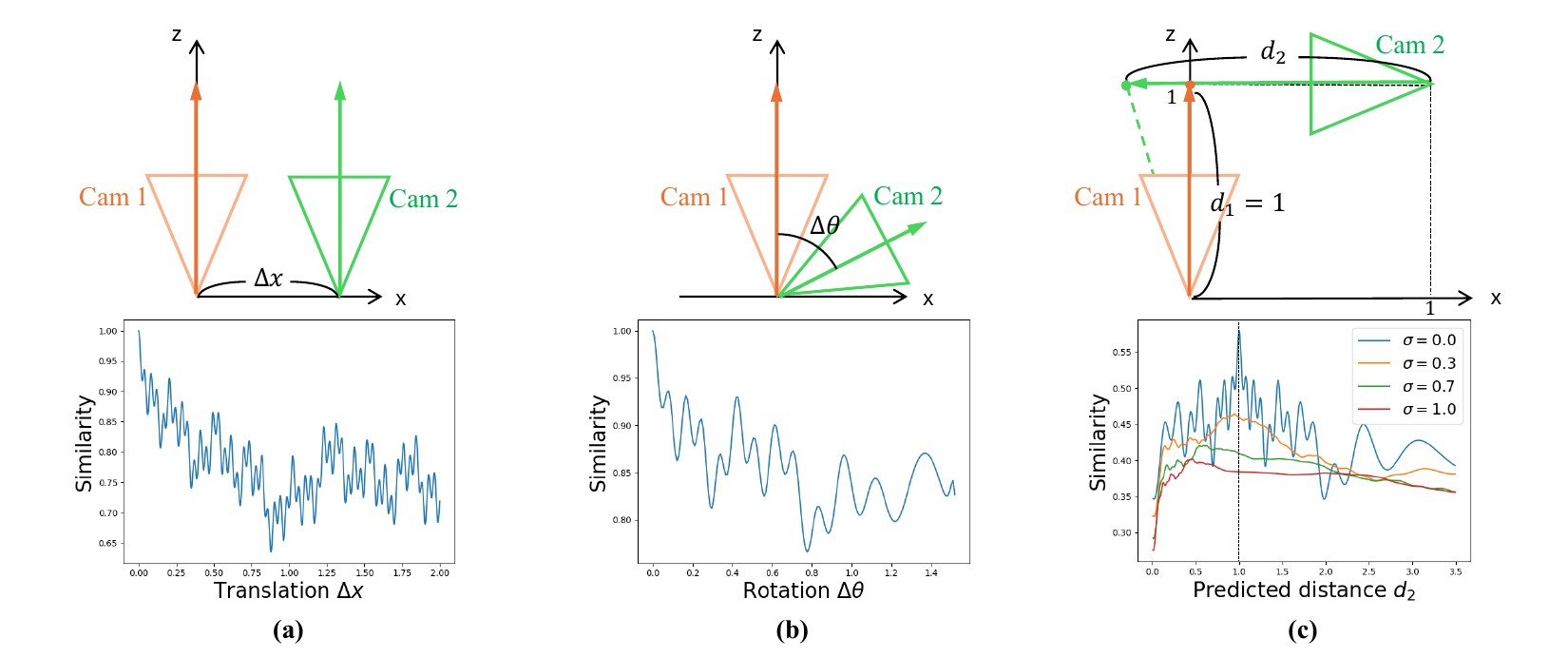}
  \caption{\textbf{Analysis of attention similarities as the ray position varies}. We plot attention similarities between two patches as their relative position varies in three settings. The top row illustrates each setting from a top-down view: \textbf{(a)} The cameras are translated by $\Delta x$. \textbf{(b)} The cameras are rotated by $\Delta \theta$. \textbf{(c)} Both patches observe the same 3D point at depth 1. We fix the predicted depth for the first patch $d_1$ to 1, while varying the predicted depth for the second patch $d_2$ and the predicted $\sigma$.}
  \label{fig:analyze-freq}
\end{figure*}
To better understand the behavior of \ours, we study the attention similarity between two image patches at the centers of two cameras under three different types of relative pose changes. We illustrate these settings in Fig.~\ref{fig:analyze-freq} and plot the attention similarity computed with \ours~as the relative pose varies. The attention similarity is given by $\mathbf{q}^\intercal_1 \, \rho(\tilde{\mathbf{x}}_1 - \tilde{\mathbf{x}}_2)\mathbf{k}_2$ (see Eq.~\ref{eq:ours-attn}). To isolate the effect of positional encoding, we manually set the query and key features to vectors of constant value 1. Higher attention similarity indicates stronger mutual attention between the patches.

In setting (a), we start with two identical cameras and translate the second camera along the $x$-axis by a varying amount $\Delta x$, while keeping the rotation, predicted depths, and uncertainties identical. Similarly, in setting (b), we rotate the second camera while keeping all other parameters fixed. In both scenarios, the attention similarity is maximized initially (when the two patches are identical). It then decreases at long range as the relative pose change increases, while oscillating at short range. The oscillations are a result of the higher-frequency channels in the multi-frequency similarities. In setting (c), we fix the poses of both cameras such that the two central patches observe the same 3D point at $(0, 0, 1)$ (depth 1). We fix the predicted depth $d_1$ for the first camera, while varying the predicted depth $d_2$ for the second camera and the uncertainty $\sigma$ for both cameras. The similarity is maximized when $d_2=1$, \ie, when the two ray segments encode the same 3D point. As $d_2$ deviates from the ground-truth depth, the similarity decreases with oscillations. This highlights that \ours~assigns higher similarity to image patches that observe the same 3D point. Furthermore, as the uncertainty $\sigma$ increases, the high-frequency oscillations are smoothed out, demonstrating how \ours~avoids unstable high-frequency encodings when uncertainties are high.

\section{Additional Results}
\label{sec:add-result}
% \noindent\textbf{Application to a Multi-view Diffusion Model.}
% We apply \ours{} to EscherNet~\cite{cape}, a large ($\sim$800M parameters) multi-view conditioned diffusion model for novel view synthesis. Unlike LVSM that applies self-attention on all views, EscherNet performs cross-attention between reference and target views. We replace the original CaPE positional encoding in EscherNet with PRoPE or \ours{}, and train/evaluate the resulting variants on DL3DV~\cite{dl3dv}. Given a varying number of reference views, the model generates three target views, and we evaluate the results using standard metrics. The results are summarized in Table~\ref{tab:eschernet} and illustrated in Fig.~\ref{fig:example-eschernet}. Across different numbers of reference views, \ours{} consistently outperforms both the original model and the PRoPE variant, with the performance gap increasing as the number of reference views grows. These results verifies the effectiveness of \ours \ when scaled to larger model and a greater number of views.

\begin{figure}[t]
  \centering
  \includegraphics[width=\columnwidth]{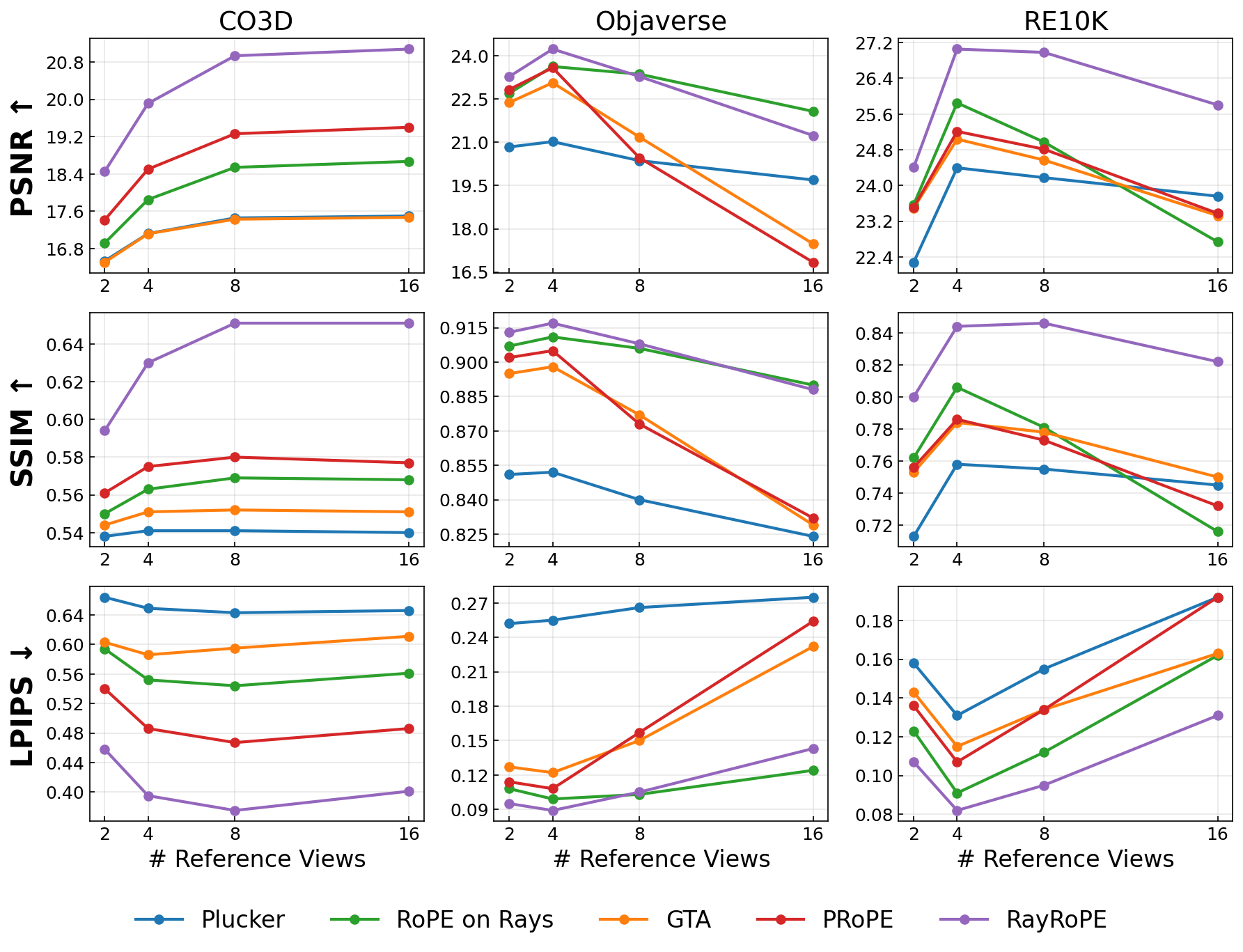}
  \caption{\textbf{Effect of varying the number of reference views.} We evaluate the LVSM model with different encodings using a varying number of reference views, while the models are trained with only 2 reference views. As the number of reference views increases, performance can increase initially (more context) but decreases later due to out-of-distribution inputs. \ours{} performs best across all settings, showing strong robustness to out-of-distribution inputs.}
  \label{fig:var-ref-views}
\end{figure}

\noindent\textbf{Generalization to varying reference views.}
As briefly introduced in Sec.~\ref{sec:exp-analysis}, we evaluate LVSM models trained with 2 reference views using a varying number of reference views at test time (zero-shot). We present the full results in Fig.~\ref{fig:var-ref-views}. Across three datasets, \ours{} outperforms all baselines in most settings. In particular, \ours{}' advantage over existing camera-based encodings (e.g., PRoPE, GTA) widens as the number of reference views increases, demonstrating stronger robustness when generalizing to unseen settings.

\noindent\textbf{Generalization to out-of-domain scenes.}
We also present full results for domain generalization in Table~\ref{tab:domain-gen}. For models trained on CO3D, we evaluate on RE10K, Objaverse, and the 10 held-out categories of CO3D. On Objaverse and RE10K, while all methods' performance degrades due to the distribution gap, \ours{} maintains its advantage over other methods. These results show that although \ours{} relies on depth prediction for geometric adaptiveness, it remains robust on out-of-distribution scenes.

\noindent\textbf{Radial vs. compound pose variations.}
We evaluate the LVSM models across different types of pose variations between reference and target views on Objaverse~\cite{objaverse}. We distinguish between \textit{radial variations}, where views share viewing angles (elevation and azimuth) but differ in intrinsics and radius to world center, and \textit{compound variations}, which introduce additional changes in viewing angles.

We report performance on these subsets in Table~\ref{tab:nvs-objv-radsph}. We observe that for targets with only radial variations, ray-based encodings (\ours~, RoPE on rays) significantly outperform camera-based baselines (GTA, PRoPE). For the radial variations only subset, the rays from reference views and target views overlap significantly. The primary synthesis challenge shifts toward the accurate reconstruction of high-frequency texture details from reference views with the same rotation. This performance gap highlights that by adopting multi-frequency encodings, \ours{} enables the attention mechanism to better reason with and transfer fine-grained details present in the reference features.

%%% EscherNet Results

\begin{figure*}
  \centering
  \includegraphics[width=\textwidth]{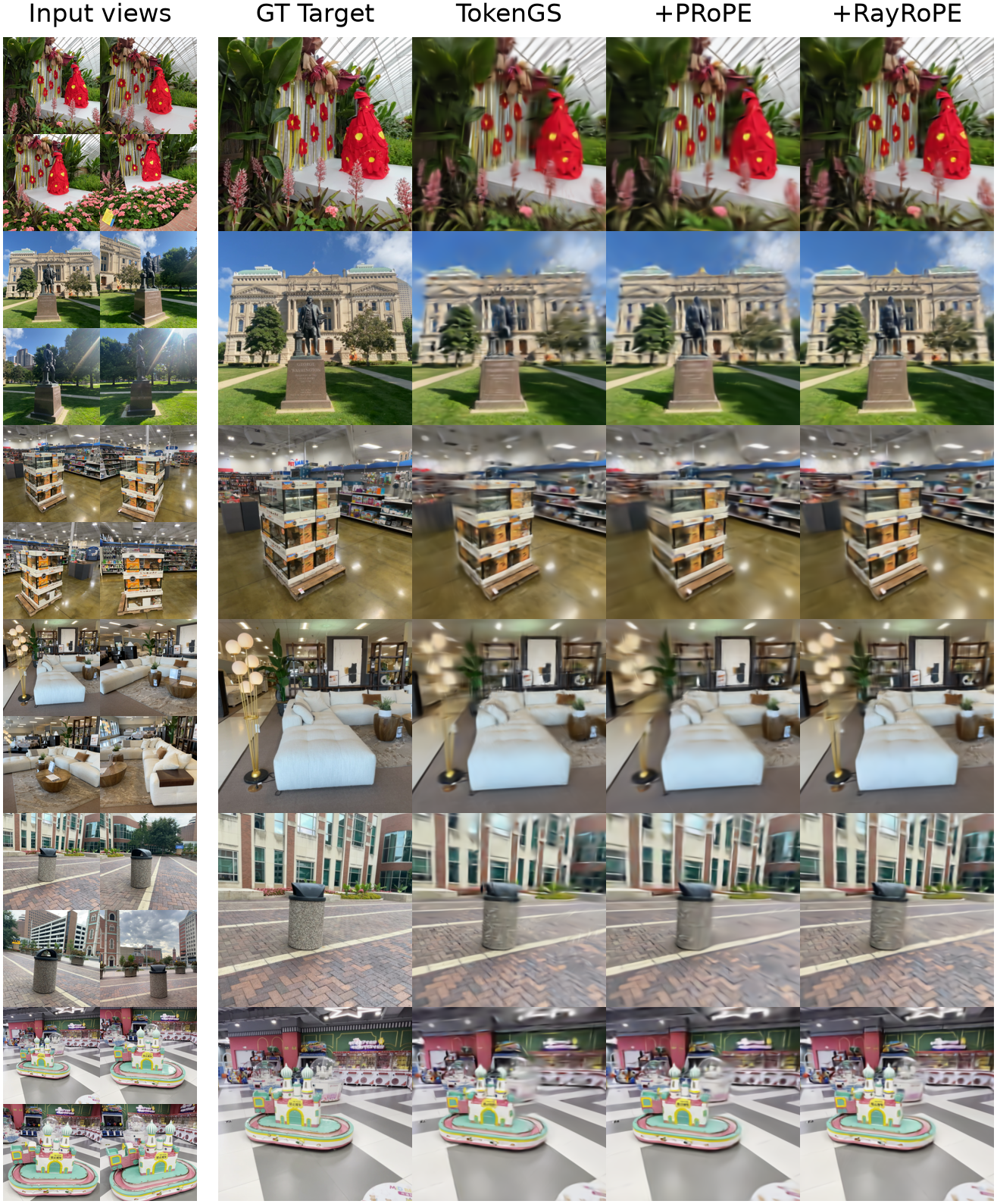}

  \caption{\textbf{Qualitative Results on TokenGS}. TokenGS takes 4 input views (left) and predicts the 3DGS for the scene. On the right, we compare renderings of the output Gaussians from different model variants. The model with \ours{} produces Gaussian splats that render images with better visual quality.}
  \label{fig:example-tokengs}
\end{figure*}

\begin{figure*}
  \centering
  \includegraphics[width=\textwidth]{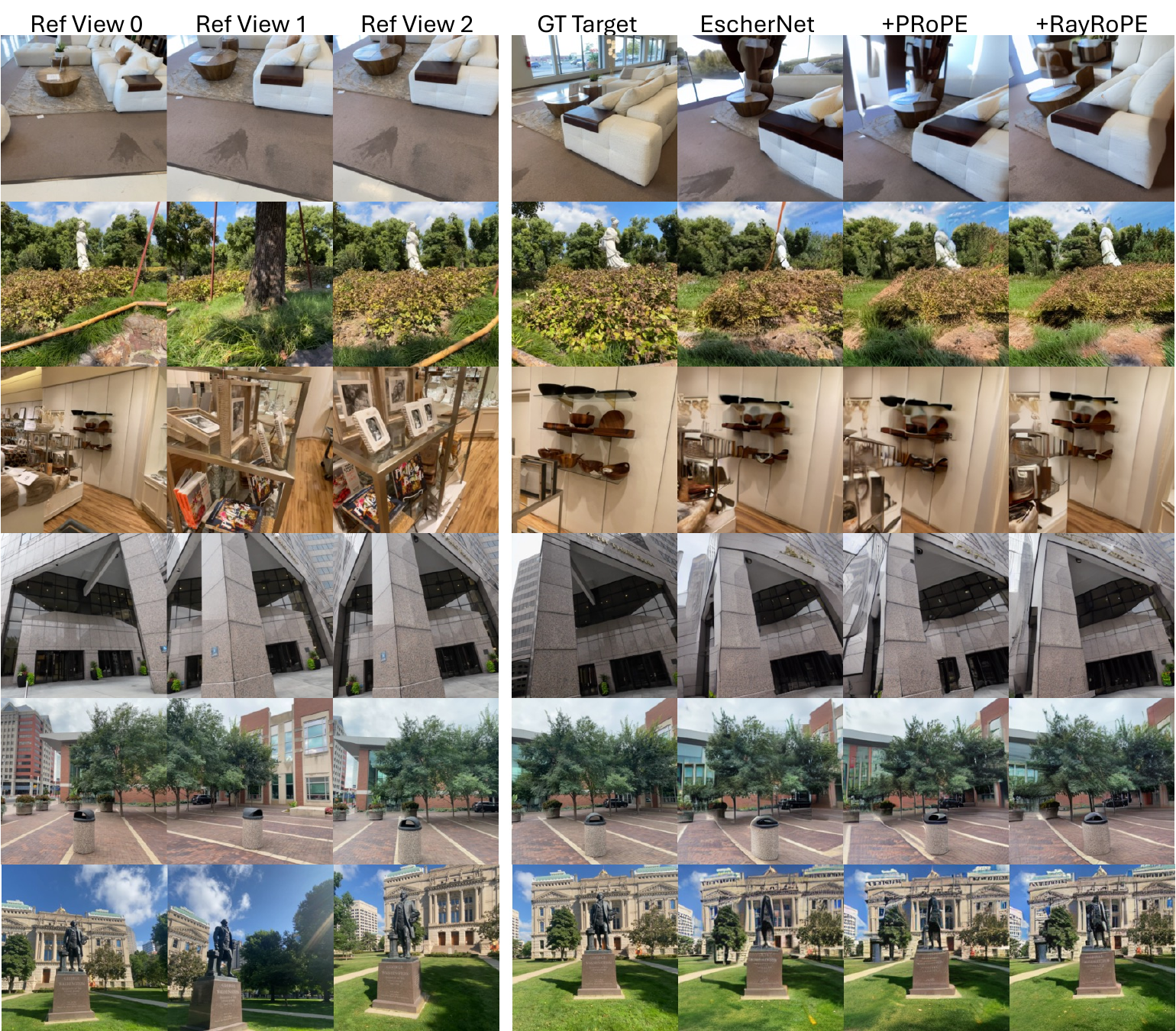}

  \caption{\textbf{Qualitative Results on EscherNet}. EscherNet conditions on 3 reference views (left) and generates target views (right) given the specified camera. With \ours{}, the model can generate target views with better visual quality and greater 3D consistency.}
  \label{fig:example-eschernet}
\end{figure*}

\clearpage

\begin{table}[t]
\centering
% \scriptsize
\tiny
\resizebox{\columnwidth}{!}{%
\begin{tabular}{l|ccc|ccc|ccc}
\toprule
\multirow{2}{*}{\raisebox{-0.8ex}{Method}}
  & \multicolumn{3}{c|}{CO3D Unseen Categories}
  & \multicolumn{3}{c|}{Objaverse}
  & \multicolumn{3}{c}{RE10K} \\
\cmidrule(lr){2-4}\cmidrule(lr){5-7}\cmidrule(lr){8-10}
 & PSNR$\uparrow$ & LPIPS$\downarrow$ & SSIM$\uparrow$
 & PSNR$\uparrow$ & LPIPS$\downarrow$ & SSIM$\uparrow$
 & PSNR$\uparrow$ & LPIPS$\downarrow$ & SSIM$\uparrow$ \\
\midrule
Plucker raymap 
 &17.08 &0.639 &0.579 &10.81 &0.523 &0.714 &15.78 &0.680 &0.493 \\
RoPE on rays 
 &17.57 &0.593 &0.560 &14.22 &0.394 &0.762 &19.00 &0.437 &0.578 \\
GTA~\cite{gta}            
 &16.99 &0.576 &0.585 &14.44 &0.342 &0.753 &18.36 &0.460 &0.568 \\
PRoPE~\cite{prope}          
 &18.28 &0.488 &0.608 &15.56 &0.327 &0.767 &19.21 &0.383 &0.588 \\
\ours        
 &\textbf{19.31} &\textbf{0.422} &\textbf{0.636} &\textbf{15.88} &\textbf{0.322} &\textbf{0.774} &\textbf{20.78} &\textbf{0.299} &\textbf{0.655} \\
\bottomrule
\end{tabular}
}
\caption{\textbf{Generalization to out-of-domain scenes.} We take the LVSM model trained on CO3D and evaluate it zero-shot on Objaverse, RE10K, and the held-out categories of CO3D. Even on out-of-domain scenes, \ours{} consistently outperforms the baseline methods.}
\label{tab:domain-gen}
\end{table}

\begin{table*}[t]
\centering
% \scriptsize
\tiny
\resizebox{0.8\textwidth}{!}{%
\begin{tabular}{l|ccc| ccc}
\toprule
\multirow{2}{*}{\raisebox{-0.8ex}{Method}}
  & \multicolumn{3}{c|}{Radial pose variations only}
  & \multicolumn{3}{c}{Compound pose variations} \\
\cmidrule(lr){2-7}
 & PSNR$\uparrow$ & LPIPS$\downarrow$ & SSIM$\uparrow$
 & PSNR$\uparrow$ & LPIPS$\downarrow$ & SSIM$\uparrow$ \\
\midrule
Plucker raymap 
 &22.76 &0.227 &0.862
 &18.93 &0.304 &0.832 \\
RoPE on rays 
 &28.50 &0.055 &0.937
 &20.47 &0.159 &0.878 \\
GTA~\cite{gta}            
 &26.71 &0.090 &0.914
 &20.39 &0.164 &0.876 \\
PRoPE~\cite{prope}          
 &27.54 &0.072 &0.924
 &20.60 &0.157 &0.879 \\
\ours        
 &\textbf{28.94} &\textbf{0.047} &\textbf{0.943}
 &\textbf{20.69} &\textbf{0.153} &\textbf{0.880} \\
\bottomrule
\end{tabular}
}
\caption{\textbf{Comparison on different types of target view changes on Objaverse~\cite{objaverse}}. The radial pose variation subset contains target views with the same viewing angle (elevation and azimuth) as one of the reference views but different intrinsics and radius. For compound pose variations, the viewing angles are also different. Multi-frequency encodings allow \ours~to significantly outperform camera-based methods (PRoPE, GTA).}
\label{tab:nvs-objv-radsph}
\end{table*}

\newpage

\begin{figure*}[t]
  \centering
  \includegraphics[width=\textwidth]{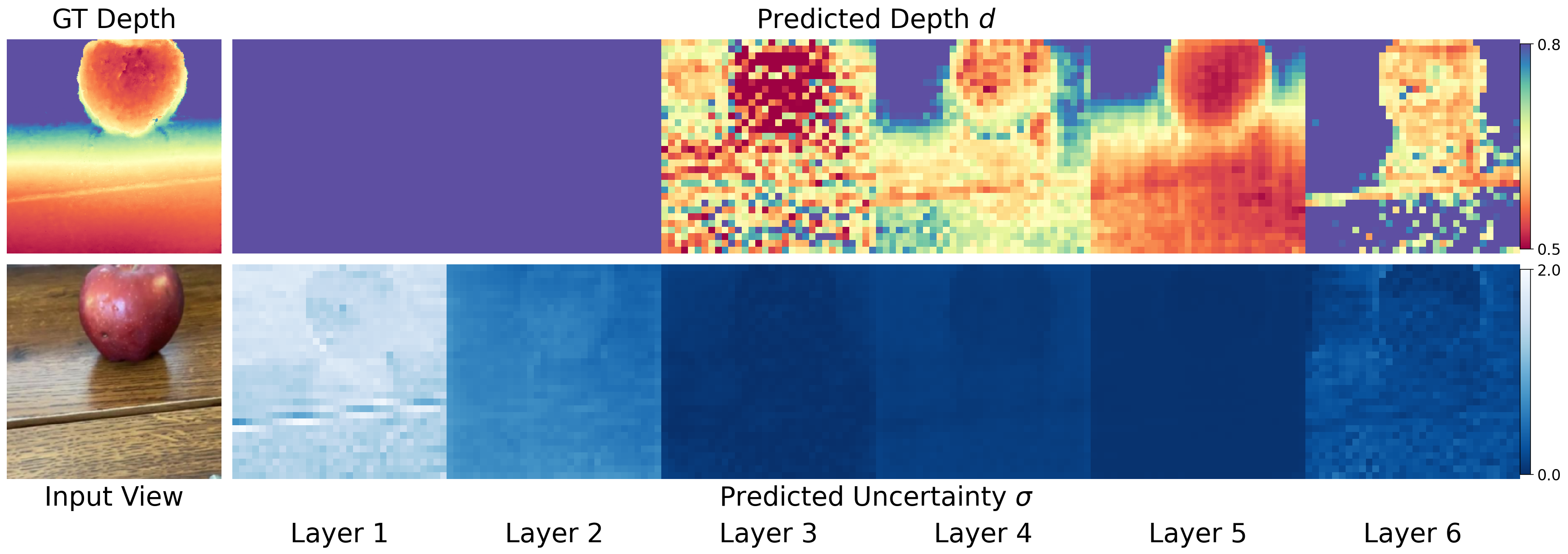}\\[0.5em]
  \includegraphics[width=\textwidth]{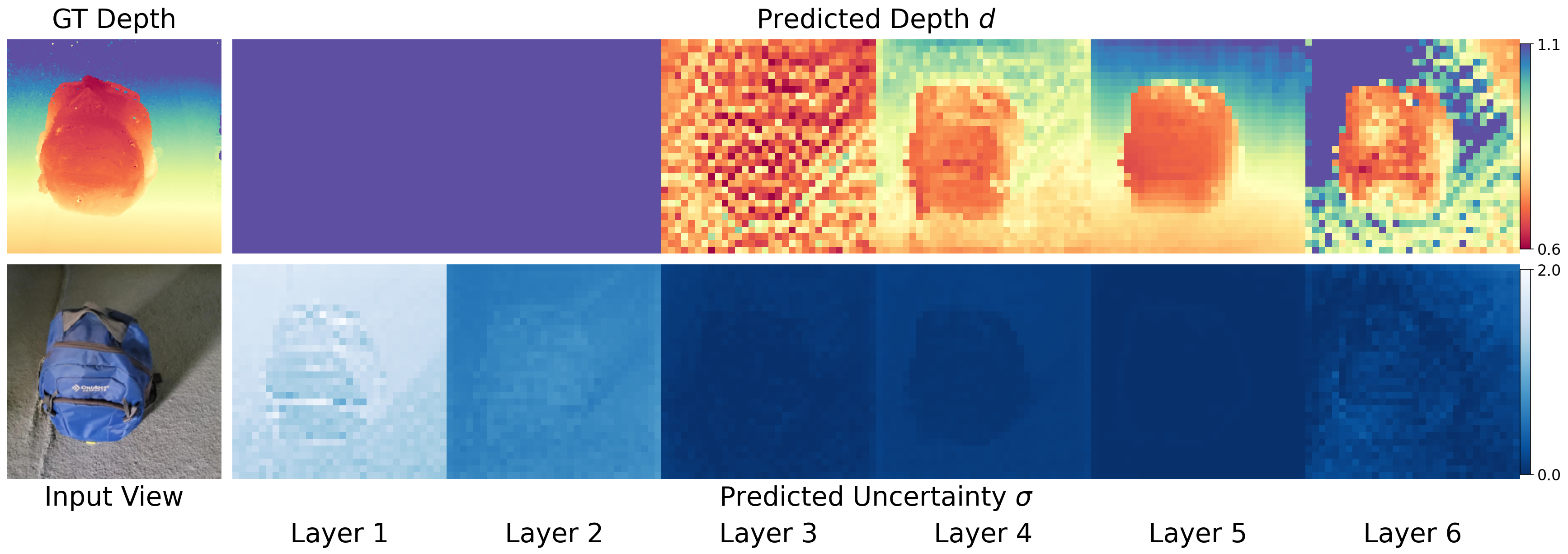}\\[0.5em]
  \includegraphics[width=\textwidth]{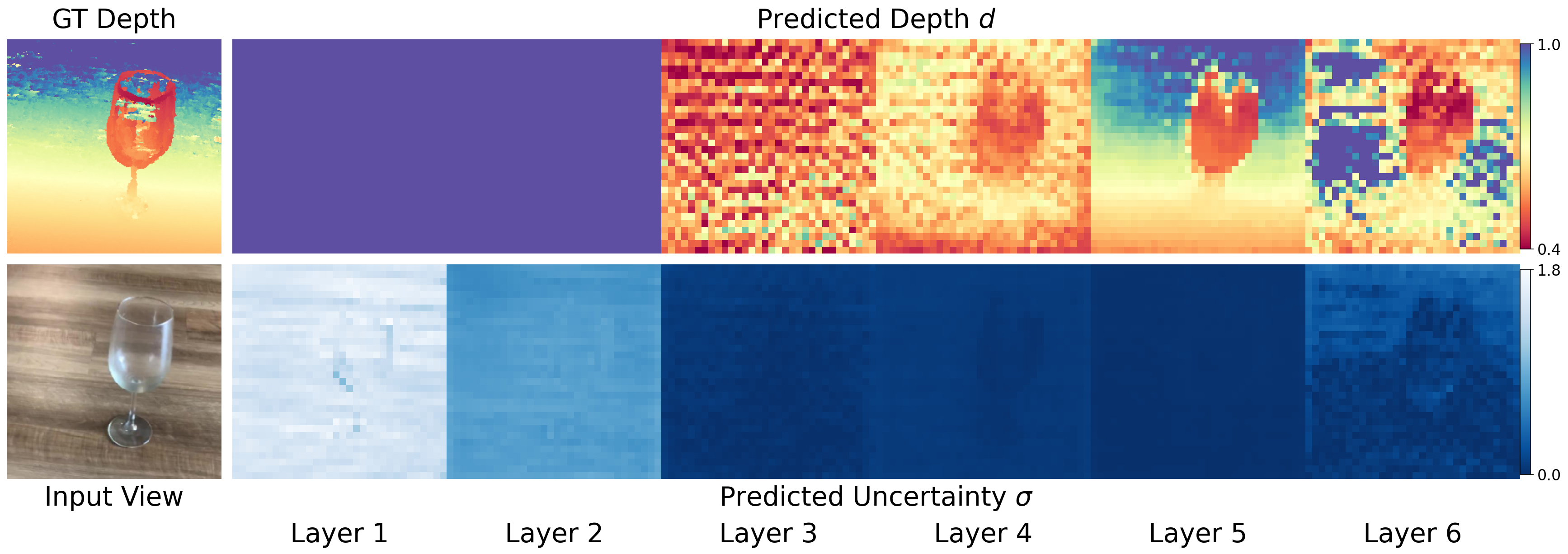}\\[0.5em]
  \includegraphics[width=\textwidth]{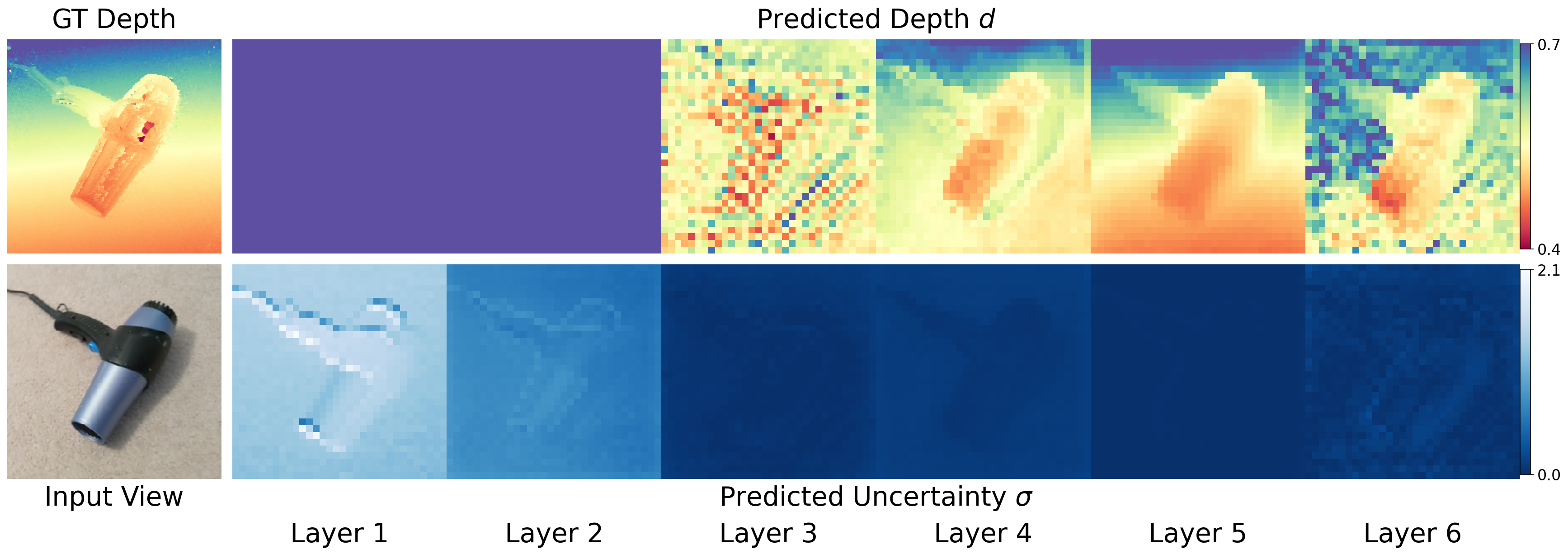}

  \caption{\textbf{Additional predicted depth visualizations}. We visualize the per-layer depth and uncertainty predictions by \ours{} on 4 additional example scenes from CO3D. Overall, the Layer 5 depth prediction matches the ground truth most closely. The predicted uncertainty is higher (lighter color) in earlier layers and lower in later layers, agreeing with our analysis in Fig.~\ref{fig:predicted_d}.}
  \label{fig:extra-pred-depths}
\end{figure*}

\begin{figure*}
  \centering
  \includegraphics[width=\textwidth]{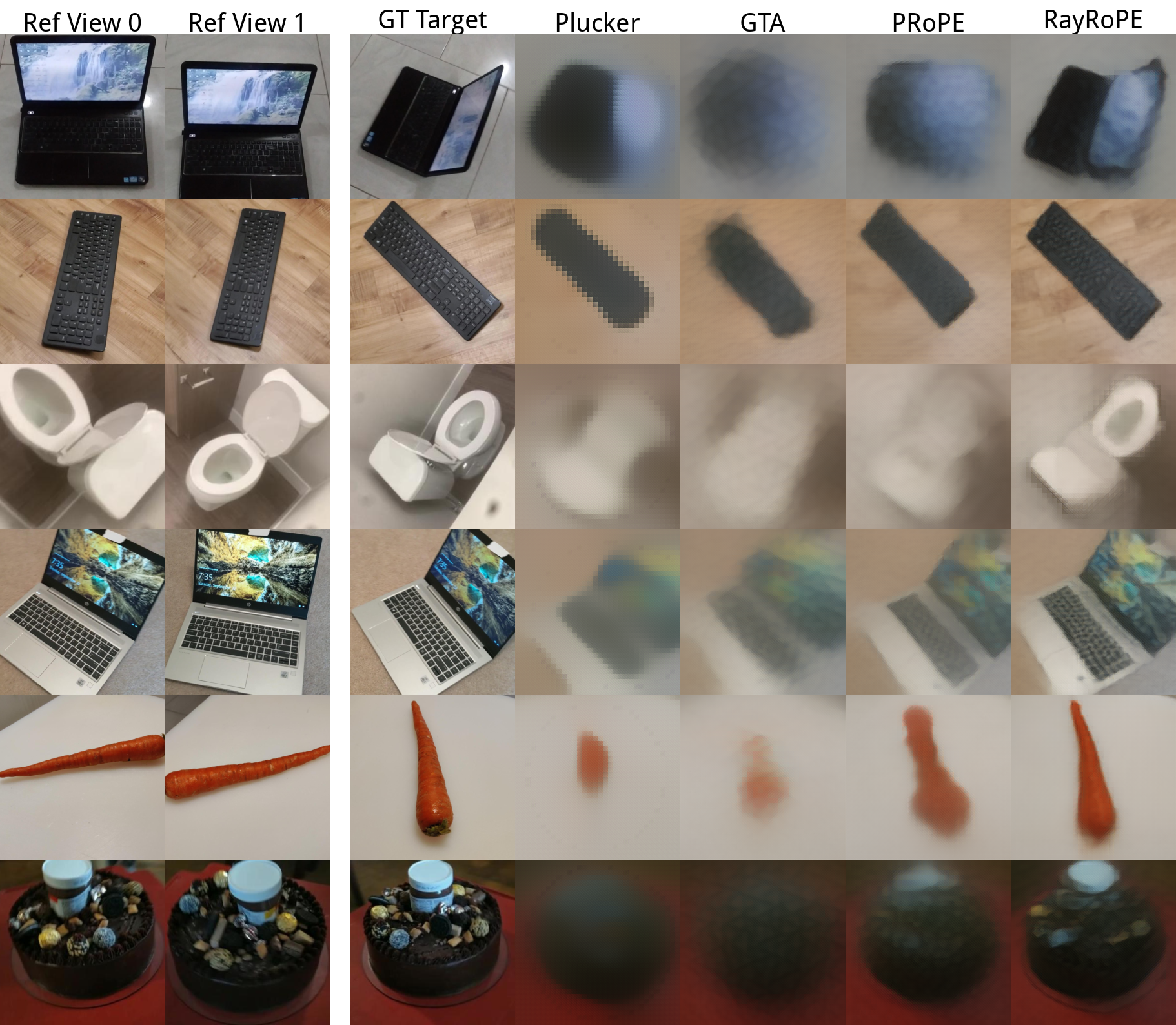}

  \caption{\textbf{Additional examples from CO3D}. Following Fig. \ref{fig:nvs-example}, we show additional qualitative comparison examples from CO3D.}
  \label{fig:example-co3d}
\end{figure*}

\begin{figure*}
  \centering
  \includegraphics[width=\textwidth]{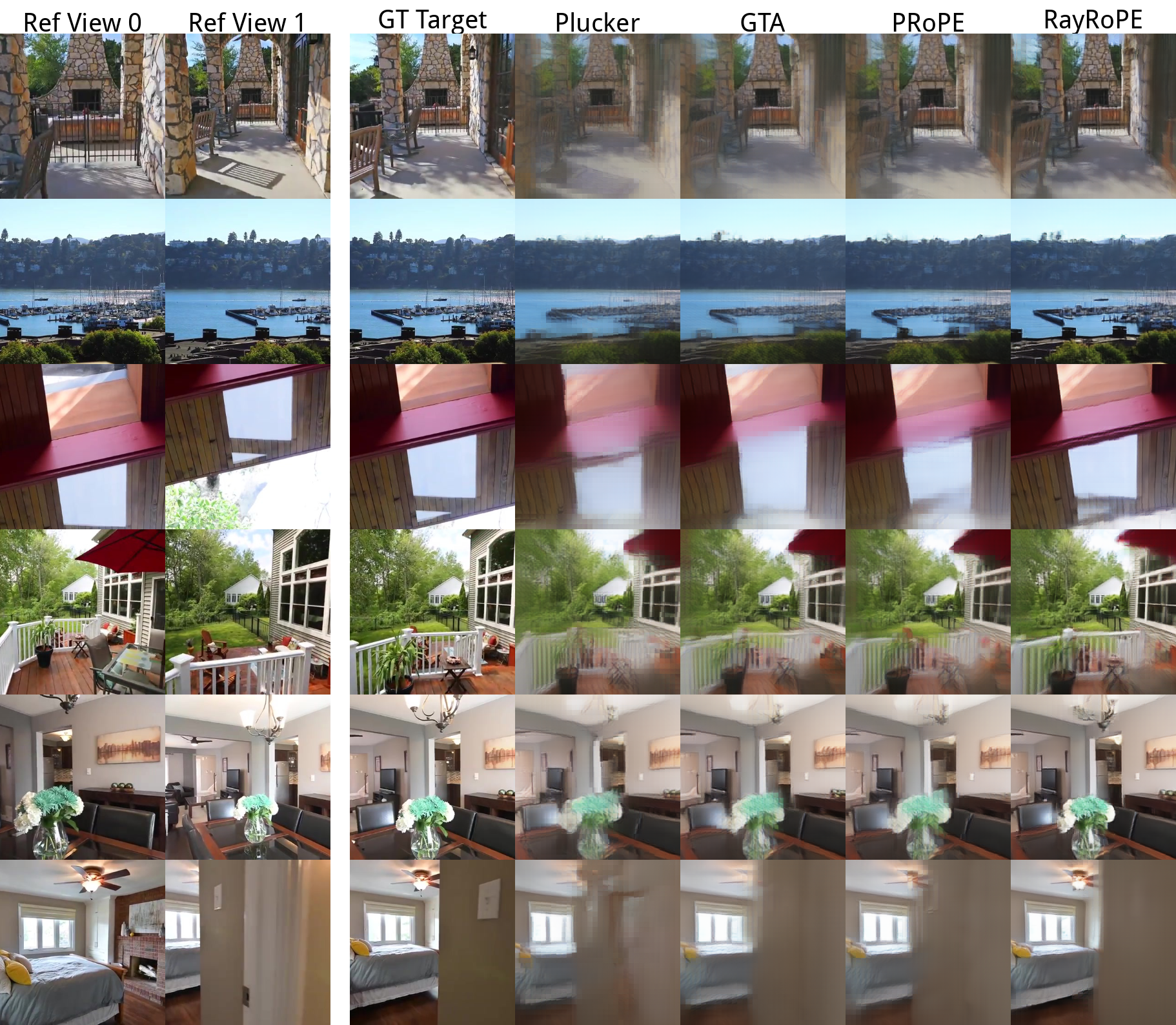}

  \caption{\textbf{Additional examples from RE10K}. Following Fig.~\ref{fig:nvs-example}, we show additional qualitative comparison examples from RE10K.}
  \label{fig:example-re10k}
\end{figure*}

\begin{figure*}
  \centering
  \includegraphics[width=\textwidth]{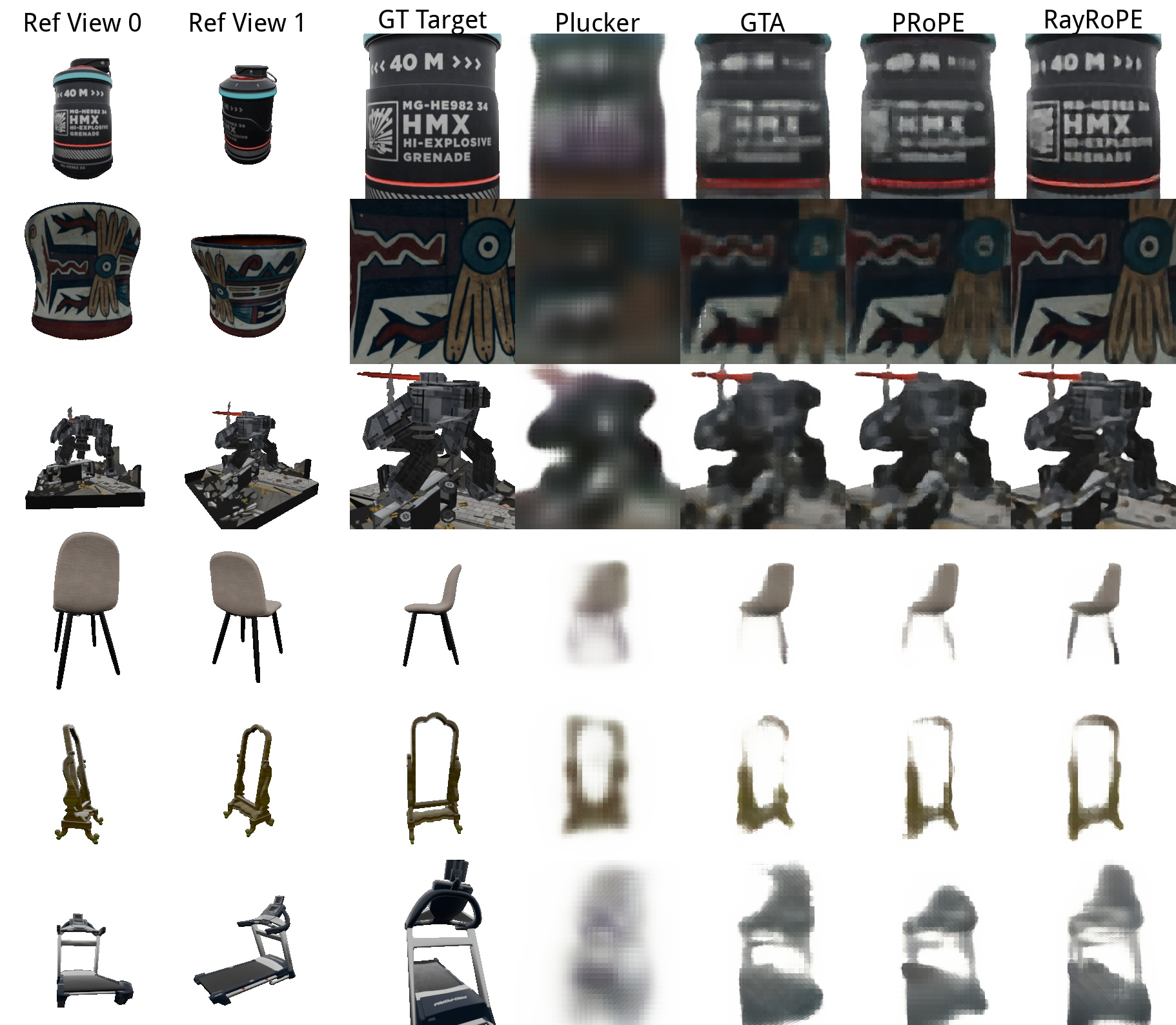}

  \caption{\textbf{Additional examples from Objaverse}. The top three rows show target views with radial variations only, while the bottom three rows show target views with compound variations.}
  \label{fig:example-objv}
\end{figure*}

% \\section{Analysis on RayRoPE}

% vs frequency\\
% vs uncertainty\\

%% file: main.bib
@String(CVPR= {IEEE Conf. Comput. Vis. Pattern Recog.})

@String(ICCV= {Int. Conf. Comput. Vis.})

@String(ECCV= {Eur. Conf. Comput. Vis.})

@String(ICLR = {Int. Conf. Learn. Represent.})

@String(CVPR  = {CVPR})

@String(ICCV  = {ICCV})

@String(ECCV  = {ECCV})

@String(ICLR  = {ICLR})

@inproceedings{re10k,
    title	= {Stereo Magnification: Learning view synthesis using multiplane images},
    author	= {Tinghui Zhou and Richard Tucker and John Flynn and Graham Fyffe and Noah Snavely},
    year	= {2018},
    booktitle	= {ACM SIGGRAPH}
}

@inproceedings{dl3dv,
  title={Dl3dv-10k: A large-scale scene dataset for deep learning-based 3d vision},
  author={Ling, Lu and Sheng, Yichen and Tu, Zhi and Zhao, Wentian and Xin, Cheng and Wan, Kun and Yu, Lantao and Guo, Qianyu and Yu, Zixun and Lu, Yawen and others},
  booktitle={CVPR},
  year={2024}
}

@inproceedings{objaverse,
  title={Objaverse: A universe of annotated 3d objects},
  author={Deitke, Matt and Schwenk, Dustin and Salvador, Jordi and Weihs, Luca and Michel, Oscar and VanderBilt, Eli and Schmidt, Ludwig and Ehsani, Kiana and Kembhavi, Aniruddha and Farhadi, Ali},
  booktitle={CVPR},
  year={2023}
}

@inproceedings{co3d,
  title={Common objects in 3d: Large-scale learning and evaluation of real-life 3d category reconstruction},
  author={Reizenstein, Jeremy and Shapovalov, Roman and Henzler, Philipp and Sbordone, Luca and Labatut, Patrick and Novotny, David},
  booktitle={ICCV},
  year={2021}
}

@inproceedings{scannet,
  title={Scannet: Richly-annotated 3d reconstructions of indoor scenes},
  author={Dai, Angela and Chang, Angel X and Savva, Manolis and Halber, Maciej and Funkhouser, Thomas and Nie{\ss}ner, Matthias},
  booktitle={CVPR},
  year={2017}
}

@InProceedings{DeMon,
  author       = "B. Ummenhofer and H. Zhou and J. Uhrig and N. Mayer and E. Ilg and A. Dosovitskiy and T. Brox",
  title        = "DeMoN: Depth and Motion Network for Learning Monocular Stereo",
  booktitle    = "CVPR",
  year         = "2017",
}

@inproceedings{prope,
  title={Cameras as relative positional encoding},
  author={Li, Ruilong and Yi, Brent and Liu, Junchen and Gao, Hang and Ma, Yi and Kanazawa, Angjoo},
  booktitle={NeurIPS},
  year={2025}
}

@inproceedings{lvsm,
  title={LVSM: A Large View Synthesis Model with Minimal 3D Inductive Bias},
  author={Jin, Haian and Jiang, Hanwen and Tan, Hao and Zhang, Kai and Bi, Sai and Zhang, Tianyuan and Luan, Fujun and Snavely, Noah and Xu, Zexiang},
  booktitle={ICLR},
  year={2025},
}

@inproceedings{unimatch,
  title={Unifying flow, stereo and depth estimation},
  author={Xu, Haofei and Zhang, Jing and Cai, Jianfei and Rezatofighi, Hamid and Yu, Fisher and Tao, Dacheng and Geiger, Andreas},
  booktitle={TPAMI},
  year={2023},
}

@inproceedings{odin,
  title={Odin: a single model for 2D and 3D segmentation},
  author={Jain, Ayush and Katara, Pushkal and Gkanatsios, Nikolaos and Harley, Adam W and Sarch, Gabriel and Aggarwal, Kriti and Chaudhary, Vishrav and Fragkiadaki, Katerina},
  booktitle={CVPR},
  year={2024}
}

@inproceedings{cat3d,
  title={CAT3D: create anything in 3D with multi-view diffusion models},
  author={Gao, Ruiqi and Ho{\l}y{\'n}ski, Aleksander and Henzler, Philipp and Brussee, Arthur and Martin-Brualla, Ricardo and Srinivasan, Pratul and Barron, Jonathan T and Poole, Ben},
  booktitle={NeurIPS},
  year={2024}
}

@inproceedings{camera_as_rays,
  title={Cameras as Rays: Pose Estimation via Ray Diffusion},
  author={Zhang, Jason Y and Lin, Amy and Kumar, Moneish and Yang, Tzu-Hsuan and Ramanan, Deva and Tulsiani, Shubham},
  booktitle={ICLR},
  year={2024}
}

@article{rope,
  title={Roformer: Enhanced transformer with rotary position embedding},
  author={Su, Jianlin and Ahmed, Murtadha and Lu, Yu and Pan, Shengfeng and Bo, Wen and Liu, Yunfeng},
  journal={Neurocomputing},
  volume={568},
  year={2024},
  publisher={Elsevier}
}

@inproceedings{cape,
  title={Eschernet: A generative model for scalable view synthesis},
  author={Kong, Xin and Liu, Shikun and Lyu, Xiaoyang and Taher, Marwan and Qi, Xiaojuan and Davison, Andrew J},
  booktitle={CVPR},
  year={2024}
}

@inproceedings{gta,
  title={GTA: A Geometry-Aware Attention Mechanism for Multi-View Transformers},
  author={Miyato, Takeru and Jaeger, Bernhard and Welling, Max and Geiger, Andreas},
  booktitle={ICLR},
  year={2024},
}

@article{kaleido,
  title={Scaling Sequence-to-Sequence Generative Neural Rendering},
  author={Liu, Shikun and Ng, Kam Woh and Jang, Wonbong and Guo, Jiadong and Han, Junlin and Liu, Haozhe and Douratsos, Yiannis and P{\'e}rez, Juan C and Zhou, Zijian and Phung, Chi and others},
  journal={arXiv preprint arXiv:2510.04236},
  year={2025}
}

@article{penc-field,
  title={Positional Encoding Field},
  author={Bai, Yunpeng and Li, Haoxiang and Huang, Qixing},
  journal={arXiv preprint arXiv:2510.20385},
  year={2025}
}

@inproceedings{bullettime,
  title={Bullettime: Decoupled control of time and camera pose for video generation},
  author={Wang, Yiming and Zhang, Qihang and Cai, Shengqu and Wu, Tong and Ackermann, Jan and Kuang, Zhengfei and Zheng, Yang and Raji{\v{c}}, Frano and Tang, Siyu and Wetzstein, Gordon},
  booktitle={CVPR},
  year={2026}
}

@article{hunyuanvideo,
  title={Hunyuanvideo: A systematic framework for large video generative models},
  author={Kong, Weijie and Tian, Qi and Zhang, Zijian and Min, Rox and Dai, Zuozhuo and Zhou, Jin and Xiong, Jiangfeng and Li, Xin and Wu, Bo and Zhang, Jianwei and others},
  journal={arXiv preprint arXiv:2412.03603},
  year={2024}
}

@article{dinov3,
  title={Dinov3},
  author={Sim{\'e}oni, Oriane and Vo, Huy V and Seitzer, Maximilian and Baldassarre, Federico and Oquab, Maxime and Jose, Cijo and Khalidov, Vasil and Szafraniec, Marc and Yi, Seungeun and Ramamonjisoa, Micha{\"e}l and others},
  journal={arXiv preprint arXiv:2508.10104},
  year={2025}
}

@inproceedings{sam2,
  title={SAM 2: Segment Anything in Images and Videos},
  author={Ravi, Nikhila and Gabeur, Valentin and Hu, Yuan-Ting and Hu, Ronghang and Ryali, Chaitanya and Ma, Tengyu and Khedr, Haitham and R{\"a}dle, Roman and Rolland, Chloe and Gustafson, Laura and others},
  booktitle={ICLR},
  year={2025}
}

@inproceedings{cogvideox,
  title={CogVideoX: Text-to-Video Diffusion Models with An Expert Transformer},
  author={Yang, Zhuoyi and Teng, Jiayan and Zheng, Wendi and Ding, Ming and Huang, Shiyu and Xu, Jiazheng and Yang, Yuanming and Hong, Wenyi and Zhang, Xiaohan and Feng, Guanyu and others},
  booktitle={ICLR},
  year={2025}
}

@inproceedings{transformer,
  title={Attention is all you need},
  author={Vaswani, Ashish and Shazeer, Noam and Parmar, Niki and Uszkoreit, Jakob and Jones, Llion and Gomez, Aidan N and Kaiser, {\L}ukasz and Polosukhin, Illia},
  booktitle={NeurIPS},
  year={2017}
}

@inproceedings{bert,
  title={Bert: Pre-training of deep bidirectional transformers for language understanding},
  author={Devlin, Jacob and Chang, Ming-Wei and Lee, Kenton and Toutanova, Kristina},
  booktitle={NAACL-HLT},
  year={2019}
}

@inproceedings{vit,
  title={An image is worth 16x16 words: Transformers for image recognition at scale},
  author={Dosovitskiy, Alexey and Beyer, Lucas and Kolesnikov, Alexander and Weissenborn, Dirk and Zhai, Xiaohua and Unterthiner, Thomas and  Dehghani, Mostafa and Minderer, Matthias and Heigold, Georg and Gelly, Sylvain and Uszkoreit, Jakob and Houlsby, Neil},
  booktitle={ICLR},
  year={2021}
}

@inproceedings{dino,
  title={Emerging properties in self-supervised vision transformers},
  author={Caron, Mathilde and Touvron, Hugo and Misra, Ishan and J{\'e}gou, Herv{\'e} and Mairal, Julien and Bojanowski, Piotr and Joulin, Armand},
  booktitle={ICCV},
  year={2021}
}

@inproceedings{clip,
  title={Learning transferable visual models from natural language supervision},
  author={Radford, Alec and Kim, Jong Wook and Hallacy, Chris and Ramesh, Aditya and Goh, Gabriel and Agarwal, Sandhini and Sastry, Girish and Askell, Amanda and Mishkin, Pamela and Clark, Jack and others},
  booktitle={ICML},
  year={2021}
}

@inproceedings{latentdiffusion,
  title={High-resolution image synthesis with latent diffusion models},
  author={Rombach, Robin and Blattmann, Andreas and Lorenz, Dominik and Esser, Patrick and Ommer, Bj{\"o}rn},
  booktitle={CVPR},
  year={2022}
}

@inproceedings{sam,
  title={Segment anything},
  author={Kirillov, Alexander and Mintun, Eric and Ravi, Nikhila and Mao, Hanzi and Rolland, Chloe and Gustafson, Laura and Xiao, Tete and Whitehead, Spencer and Berg, Alexander C and Lo, Wan-Yen and others},
  booktitle={ICCV},
  year={2023}
}

@inproceedings{tttransformer,
  title={Exploring the limits of transfer learning with a unified text-to-text transformer},
  author={Raffel, Colin and Shazeer, Noam and Roberts, Adam and Lee, Katherine and Narang, Sharan and Matena, Michael and Zhou, Yanqi and Li, Wei and Liu, Peter J},
  booktitle={JMLR},
  year={2020}
}

@article{llama,
  title={Llama: Open and efficient foundation language models},
  author={Touvron, Hugo and Lavril, Thibaut and Izacard, Gautier and Martinet, Xavier and Lachaux, Marie-Anne and Lacroix, Timoth{\'e}e and Rozi{\`e}re, Baptiste and Goyal, Naman and Hambro, Eric and Azhar, Faisal and others},
  journal={arXiv preprint arXiv:2302.13971},
  year={2023}
}

@article{deepseek,
  title={Deepseek-r1: Incentivizing reasoning capability in llms via reinforcement learning},
  author={Guo, Daya and Yang, Dejian and Zhang, Haowei and Song, Junxiao and Zhang, Ruoyu and Xu, Runxin and Zhu, Qihao and Ma, Shirong and Wang, Peiyi and Bi, Xiao and others},
  journal={arXiv preprint arXiv:2501.12948},
  year={2025}
}

@inproceedings{visualinstruction,
  title={Visual instruction tuning},
  author={Liu, Haotian and Li, Chunyuan and Wu, Qingyang and Lee, Yong Jae},
  booktitle={NeurIPS},
  year={2023}
}

@inproceedings{relpose,
  title={Self-attention with relative position representations},
  author={Shaw, Peter and Uszkoreit, Jakob and Vaswani, Ashish},
  booktitle={NAACL},
  year={2018}
}

@inproceedings{trainshort,
  title={Train short, test long: Attention with linear biases enables input length extrapolation},
  author={Press, Ofir and Smith, Noah A and Lewis, Mike},
  booktitle={ICLR},
  year={2022}
}

@inproceedings{3dconcept,
  title={3d concept learning and reasoning from multi-view images},
  author={Hong, Yining and Lin, Chunru and Du, Yilun and Chen, Zhenfang and Tenenbaum, Joshua B and Gan, Chuang},
  booktitle={CVPR},
  year={2023}
}

@inproceedings{fourier,
  title={Learnable fourier features for multi-dimensional spatial positional encoding},
  author={Li, Yang and Si, Si and Li, Gang and Hsieh, Cho-Jui and Bengio, Samy},
  booktitle={NeurIPS},
  year={2021}
}

@inproceedings{string,
  title={Learning the RoPEs: Better 2D and 3D Position Encodings with STRING},
  author={Schenck, Connor and Reid, Isaac and Jacob, Mithun George and Bewley, Alex and Ainslie, Joshua and Rendleman, David and Jain, Deepali and Sharma, Mohit and Dubey, Kumar Avinava and Wahid, Ayzaan and others},
  booktitle={ICML},
  year={2025}
}

@inproceedings{lrm,
  title={LRM: Large Reconstruction Model for Single Image to 3D},
  author={Hong, Yicong and Zhang, Kai and Gu, Jiuxiang and Bi, Sai and Zhou, Yang and Liu, Difan and Liu, Feng and Sunkavalli, Kalyan and Bui, Trung and Tan, Hao},
  booktitle={ICLR},
  year={2024}
}

@inproceedings{gs-lrm,
  title={GS-LRM: Large Reconstruction Model for 3D Gaussian Splatting},
  author={Zhang, Kai and Bi, Sai and Tan, Hao and Xiangli, Yuanbo and Zhao, Nanxuan and Sunkavalli, Kalyan and Xu, Zexiang},
  booktitle={ECCV},
  year={2024}
}

@inproceedings{seva,
  title={Stable virtual camera: Generative view synthesis with diffusion models},
  author={Zhou, Jensen and Gao, Hang and Voleti, Vikram and Vasishta, Aaryaman and Yao, Chun-Han and Boss, Mark and Torr, Philip and Rupprecht, Christian and Jampani, Varun},
  booktitle={ICCV},
  year={2025}
}

@inproceedings{bolt3d,
  title={Bolt3d: Generating 3d scenes in seconds},
  author={Szymanowicz, Stanislaw and Zhang, Jason Y and Srinivasan, Pratul and Gao, Ruiqi and Brussee, Arthur and Holynski, Aleksander and Martin-Brualla, Ricardo and Barron, Jonathan T and Henzler, Philipp},
  booktitle={ICCV},
  year={2025}
}

@inproceedings{ac3d,
  title={Ac3d: Analyzing and improving 3d camera control in video diffusion transformers},
  author={Bahmani, Sherwin and Skorokhodov, Ivan and Qian, Guocheng and Siarohin, Aliaksandr and Menapace, Willi and Tagliasacchi, Andrea and Lindell, David B and Tulyakov, Sergey},
  booktitle={CVPR},
  year={2025}
}

@inproceedings{zero123,
  title={Zero-1-to-3: Zero-shot one image to 3d object},
  author={Liu, Ruoshi and Wu, Rundi and Van Hoorick, Basile and Tokmakov, Pavel and Zakharov, Sergey and Vondrick, Carl},
  booktitle={ICCV},
  year={2023}
}

@inproceedings{2drope,
  title={Rotary position embedding for vision transformer},
  author={Heo, Byeongho and Park, Song and Han, Dongyoon and Yun, Sangdoo},
  booktitle={ECCV},
  year={2024}
}

@inproceedings{pixelsplat,
      title={pixelSplat: 3D Gaussian Splats from Image Pairs for Scalable Generalizable 3D Reconstruction},
      author={David Charatan and Sizhe Li and Andrea Tagliasacchi and Vincent Sitzmann},
      year={2024},
      booktitle={CVPR},
}

@inproceedings{lgm,
  title={LGM: Large Multi-view Gaussian Model for High-Resolution 3D Content Creation},
  author={Tang, Jiaxiang and Chen, Zhaoxi and Chen, Xiaokang and Wang, Tengfei and Zeng, Gang and Liu, Ziwei},
  booktitle={ECCV},
  year={2024},
}

@inproceedings{rgbd,
  title={A benchmark for the evaluation of RGB-D SLAM systems},
  author={Sturm, J{\"u}rgen and Engelhard, Nikolas and Endres, Felix and Burgard, Wolfram and Cremers, Daniel},
  booktitle={IROS},
  year={2012}
}

@inproceedings{sun3d,
  title={Sun3d: A database of big spaces reconstructed using sfm and object labels},
  author={Xiao, Jianxiong and Owens, Andrew and Torralba, Antonio},
  booktitle={CVPR},
  year={2013}
}

@article{flashattention,
  title={Flashattention: Fast and memory-efficient exact attention with io-awareness},
  author={Dao, Tri and Fu, Dan and Ermon, Stefano and Rudra, Atri and R{\'e}, Christopher},
  journal={Neurips},
  year={2022}
}

@article{pope2023efficiently-kvcache,
  title={Efficiently scaling transformer inference},
  author={Pope, Reiner and Douglas, Sholto and Chowdhery, Aakanksha and Devlin, Jacob and Bradbury, James and Heek, Jonathan and Xiao, Kefan and Agrawal, Shivani and Dean, Jeff},
  journal={Proceedings of machine learning and systems},
  year={2023}
}

@inproceedings{zerodepth,
  title={Towards zero-shot scale-aware monocular depth estimation},
  author={Guizilini, Vitor and Vasiljevic, Igor and Chen, Dian and Ambruș, Rareș and Gaidon, Adrien},
  booktitle={ICCV},
  year={2023}
}

@inproceedings{input-3d-bias,
  title={Input-level inductive biases for 3D reconstruction},
  author={Yifan, Wang and Doersch, Carl and Arandjelovi{\'c}, Relja and Carreira, Joao and Zisserman, Andrew},
  booktitle={CVPR},
  year={2022}
}

@article{rerope,
  title={Rerope: Repurposing rope for relative camera control},
  author={Li, Chunyang and Yang, Yuanbo and Shao, Jiahao and Zhou, Hongyu and Schwarz, Katja and Liao, Yiyi},
  journal={arXiv preprint},
  year={2026}
}

@article{viewrope,
  title={Geometry-aware rotary position embedding for consistent video world model},
  author={Xiang, Chendong and Liu, Jiajun and Zhang, Jintao and Yang, Xiao and Fang, Zhengwei and Wang, Shizun and Wang, Zijun and Zou, Yingtian and Su, Hang and Zhu, Jun},
  journal={arXiv preprint},
  year={2026}
}

@inproceedings{tokengs,
  title={Tokengs: Decoupling 3d gaussian prediction from pixels with learnable tokens},
  author={Ren, Jiawei and Tyszkiewicz, Michal Jan and Huang, Jiahui and Gojcic, Zan},
  booktitle={CVPR},
  year={2026}
}

@article{3dgs,
  title={3D Gaussian Splatting for Real-Time Radiance Field Rendering},
  author={Kerbl, Bernhard and Kopanas, Georgios and Leimk{\"u}hler, Thomas and Drettakis, George},
  journal={ACM Transactions on Graphics},
  volume={42},
  number={4},
  pages={1--14},
  year={2023}
}
